\documentclass[lettersize,journal]{IEEEtran}
\usepackage{amsmath,amsfonts}
\usepackage{algorithmic}
\usepackage{algorithm}
\usepackage{array}
\usepackage[caption=false,font=normalsize,labelfont=sf,textfont=sf]{subfig}
\usepackage{textcomp}
\usepackage{stfloats}
\usepackage{url}
\usepackage{verbatim}
\usepackage{graphicx}
\usepackage{cite}
\usepackage{hyperref}
\usepackage{cleveref}
\usepackage{xcolor}
\usepackage{fontspec}
\usepackage{pifont}
\usepackage{multirow}
\usepackage{amssymb}
\usepackage{amsthm}
\usepackage{booktabs}
\begin{document}

\title{Multivariate Time Series Anomaly Detection via Dual-Branch Reconstruction and Autoregressive Flow-based Residual Density Estimation}

% \author{IEEE Publication Technology,~\IEEEmembership{Staff,~IEEE,}
%         % <-this % stops a space
% % \thanks{This paper was produced by the IEEE Publication Technology Group. They are in Piscataway, NJ.}% <-this % stops a space
% % \thanks{Manuscript received April 19, 2021; revised August 16, 2021.}}
% }
\author{
  Jun Liu$^1$ \and
  Ying Chen$^1$ \and Ziqian Lu$^2$ \and Qinyue Tong$^1$ \and Jun Tang$^1$
  \\
  $^1$ Zhejiang University, Hangzhou, China \\
%   \\
  $^2$ Zhejiang Sci-Tech University, Hangzhou, China
%   \\
%   $^1$ junliu0930@zju.edu.cn, $^2$ yizhang@tsinghua.edu.cn
}

% The paper headers
\markboth{Journal of \LaTeX\ Class Files,~Vol.~14, No.~8, August~2021}%
{Shell \MakeLowercase{\textit{et al.}}: A Sample Article Using IEEEtran.cls for IEEE Journals}

% \IEEEpubid{0000--0000/00\$00.00~\copyright~2021 IEEE}
% Remember, if you use this you must call \IEEEpubidadjcol in the second
% column for its text to clear the IEEEpubid mark.

\maketitle

\begin{abstract}
Multivariate Time Series Anomaly Detection (MTSAD) is critical for real-world monitoring scenarios such as industrial control and aerospace systems. Mainstream reconstruction-based anomaly detection methods suffer from two key limitations: first, overfitting to spurious correlations induced by an overemphasis on cross-variable modeling; second, the generation of misleading anomaly scores by simply summing up multivariable reconstruction errors, which makes it difficult to distinguish between hard-to-reconstruct samples and genuine anomalies. To address these issues, we propose DBR-AF, a novel framework that integrates a dual-branch reconstruction (DBR) encoder and an autoregressive flow (AF) module. The DBR encoder decouples cross-variable correlation learning and intra-variable statistical property modeling to mitigate spurious correlations, while the AF module employs multiple stacked reversible transformations to model the complex multivariate residual distribution and further leverages density estimation to accurately identify normal samples with large reconstruction errors. Extensive experiments on seven benchmark datasets demonstrate that DBR-AF achieves state-of-the-art performance, with ablation studies validating the indispensability of its core components.
\end{abstract}


\section{Introduction}
\begin{figure}
    \begin{center}
        \includegraphics[width=1\linewidth]{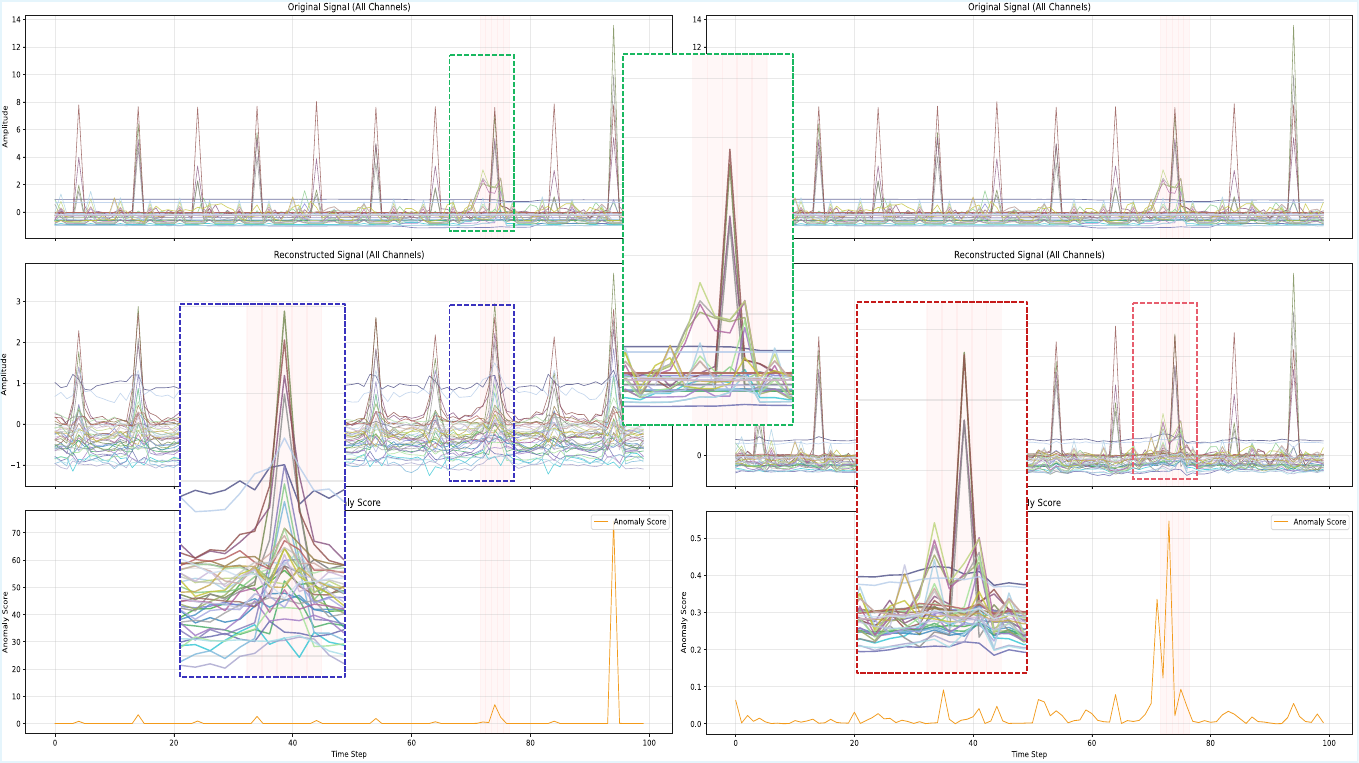}
    \end{center}
    \caption{The \textbf{left} subplot reveals the fatal flaws of existing reconstruction-based methods: cross-variable interference induces large amplitude errors and spurious non-stationary characteristics in reconstructed signals, and anomaly scores are dominated by variables with significant reconstruction errors, thereby giving rise to false positive detections. In contrast, the \textbf{right} subplot demonstrates the advantages of our proposed method: it effectively suppresses cross-variable interference to achieve high-fidelity reconstructions that align with the original multivariate time series and exhibits strong robustness to large-amplitude reconstruction errors during anomaly discrimination. Ground-truth anomaly segments are shaded in pink.}
\label{fig:1}
\end{figure}
\IEEEPARstart{G}{iven} the widespread presence of time series data in real-world applications such as industrial monitoring and aerospace systems~\cite{hundman2018detecting, rehbach2018gecco}, and its crucial role in ensuring the normal operation of equipment and environments, deep learning-based MTSAD has become a key research focus. This approach effectively addresses the growing need for automated anomaly detection in large-scale, sequential data streams~\cite{el2023multivariate, wang2025survey}. Existing MTSAD paradigms generally comprise two core modules: a representation learning layer, which extracts inter-variable correlations and temporal patterns from raw multi-channel signals to derive informative latent variables, and a decision-making layer, which formulates decision rules based on these latent features to discriminate between normal and abnormal instances. Among these paradigms, reconstruction-based methods have garnered substantial attention for their unsupervised adaptability: they are trained exclusively on unlabeled normal data to learn the intrinsic reconstruction patterns of normal time series and identify anomalies by detecting prominent reconstruction discrepancies in unseen anomalous data. For this category of methods, the representation layer commonly adopts autoencoders integrated with the Transformer's parallel attention mechanism to efficiently capture cross-variable correlations and temporal dynamics, while the decision layer relies on reconstruction errors (e.g., mean squared error, MSE) between model outputs and raw inputs for anomaly scoring~\cite{xu2021anomaly,zhang2022tfad, yang2023dcdetector, fang2024temporal, shen2025learn}.

\IEEEpubidadjcol

However, two core problems remain unresolved in existing reconstruction-based approaches, as illustrated in ~\cref{fig:1}. First, these methods overemphasize the modeling of cross-variable relationships, a tendency that often leads to the overfitting of spurious correlations~\cite{wu2022timesnet, wu2024catch, xie2025multivariate}. Such correlations are non-causal associations, stemming not from the system's intrinsic dynamics but from noise or coincidental covariation. Learning these extraneous correlation patterns compels inherently heterogeneous variables to yield homogeneous reconstruction results, which in turn exacerbates overall reconstruction errors. Furthermore, certain signals exhibit prominent periodic peaks during normal operation, and modeling such signals is inherently challenging due to their large amplitude fluctuations and non-stationary dynamics. Additional interference from stable, low-magnitude stationary signals further amplifies this modeling difficulty, ultimately leading to a significant increase in reconstruction errors for these peak-containing signals.
Second, most of these methods employ a simplistic aggregation strategy: they directly sum the reconstruction errors of individual variables to generate the final anomaly score, failing to account for the differential importance of each variable in anomaly detection. As a result, anomaly scores are frequently influenced by large-magnitude reconstruction errors arising from signals that, while inherently normal, are challenging to reconstruct with high accuracy. Even though the peak fluctuations of these signals remain within the system's normal operating range, their inherent complexity leads to inflated reconstruction errors. Consequently, the model tends to misclassify these normal fluctuations as anomalies, significantly compromising the reliability of detection outcomes.

To address the above issues, we propose a novel multivariate time series anomaly detection framework, namely, Dual-Branch Reconstruction and Autoregressive Flow-based Residual Density Estimation (DBR-AF).

Specifically, we employ a dual-branch reconstruction (DBR) encoder designed to independently learn cross-variable correlations and intra-variable statistical properties at the representation level. Among the two branches, the temporal branch focuses on capturing correlations between variables and their time-varying dynamics, while the channel branch is dedicated to modeling the statistical properties of individual variables and their inherent time-dependent behaviors. This decoupled design mitigates the interference between inter-variable and intra-variable feature learning, reducing the risk of overlearning spurious correlations.
Subsequently, we use the residuals between the original inputs and the reconstructed outputs of the DBR encoder as the target for density estimation. These residuals inherently eliminate redundant temporal dependencies, providing a more accurate representation of the error magnitude and direction for each variable. To achieve accurate density estimation of residuals, we introduce an autoregressive flow module (AF)~\cite{rezende2015variational}, which learns a series of reversible transformations that map the residual distribution to a pre-specified simple prior (e.g., a diagonal mean multivariate distribution). Notably, these stacked transformations allow the AF to approximate highly complex mappings, overcoming the limitations of rigid parametric density estimation methods~\cite{draxler2024universality}. This flexibility enables the model to capture the nuanced characteristics of multivariate residual distributions while preserving the invertibility needed for accurate likelihood estimation.
Crucially, for normally behaving variables that are inherently difficult to reconstruct (e.g., those with periodic peak fluctuations), the AF learns the specific density relationships associated with their residuals during training. Instead of treating their large but normal reconstruction errors as indicators of anomalies, the flow captures these patterns in the reversible transformation. As a result, the residuals align with the pre-specified prior distribution after transformation, yielding reasonable log-likelihood values. This approach effectively prevents false positives by distinguishing between ``modeling difficult'' and true ``anomalies''.
% It should be noted that in existing methods, the inflated reconstruction errors of such hard-to-reconstruct but normal signals often dominate the final anomaly score. This, in turn, leads the model to misclassify such normal fluctuations as anomalies, seriously undermining the reliability of detection results.
Ultimately, the log-likelihood from the AF and the reconstruction errors from the DBR encoder are used together in the decision layer as criteria for anomaly detection, leveraging complementary information to improve detection reliability.

To summarize, our main contributions are as follows:
\begin{itemize}
    \item We introduce a dual-branch encoder designed to separate cross-variable correlations from intra-variable statistical property learning. This approach effectively reduces the risk of spurious correlation overfitting in multivariate time-series anomaly detection.
    \item We utilize an autoregressive flow module for stacked reversible residual transformations that convert complex residual distributions into a simpler prior. This approach is effective in adapting to intricate reconstruction scenarios and helps to prevent the misclassification of "hard-to-reconstruct but normal" variables, thereby improving detection reliability.
    \item Comprehensive experiments on multiple public benchmarks demonstrate that DBR-AF achieves or outperforms state-of-the-art reconstruction-based methods across core metrics, verifying its effectiveness in multivariate time-series anomaly detection.
\end{itemize}

\section{Related Works}
\subsection{Time Series Anomaly Detection}
The development of time series anomaly detection technology stems from the urgent demand for monitoring equipment operating status in industrial production, particularly in manufacturing, energy, and aerospace. Traditional machine learning methods rely primarily on simple statistical information for clustering or density estimation to identify anomalies~\cite{hawkins1980identification, breunig2000lof}. The advent of deep learning has led to methods that leverage networks such as Recurrent Neural Networks (RNNs)~\cite{jordan1997serial}, Long Short-Term Memory (LSTM) networks~\cite{6795963}, and Temporal Convolutional Networks (TCNs)~\cite{lea2017temporal}, combined with autoencoder (AE)~\cite{rumelhart19881986} or variational autoencoder (VAE)~\cite{kingma2013auto} architectures, to extract latent representations of time-series data. These methods detect anomalies based on latent variable distributions~\cite{yairi2017data, zong2018deep} or reconstructed data errors~\cite{sakurada2014anomaly, park2018multimodal, su2019robust, li2021multivariate}. In recent years, efficient parallel attention mechanisms~\cite{vaswani2017attention} have been widely applied to time-series signal reconstruction. Attention-based methods directly adopt Transformers to reconstruct continuous time-series segments, using reconstruction error as the primary evaluation metric~\cite{xu2021anomaly}. Within this reconstruction-based framework, numerous impactful improvements have been proposed. AnomalyTransformer~\cite{xu2021anomaly} introduces association discrepancy and adopts a minimax strategy to enhance the discriminability of discrepancy between normal and abnormal data. DCdetector~\cite{yang2023dcdetector} proposes multi-view contrastive learning for time-series anomaly detection. TFAD~\cite{zhang2022tfad}, Dual-TF~\cite{nam2024breaking}, TFMAE~\cite{fang2024temporal}, and MtsCID~\cite{xie2025multivariate} model normal signals across multiple time-frequency domains. MEMTO~\cite{song2023memto} leverages memory modules to store normal patterns, thereby alleviating model overfitting to abnormal data. CATCH~\cite{wu2024catch} focuses on fine-grained frequency features and models multi-channel correlations. D\(^3\)R~\cite{wang2023drift} proposes dynamic decomposition and diffusion reconstruction to address time-series distribution shift, while EDAD~\cite{zhang2025encode} decomposes latent representations into stable and auxiliary features to reduce autoencoder sensitivity to abnormal training samples.

\subsection{Normalizing Flow}
The essence of normalizing flow lies in transforming a simple distribution into a complex one via invertible transformations~\cite{dinh2014nice}. Rezende et al.~\cite{rezende2015variational} incorporated normalizing flows into variational inference, demonstrating that stacking multiple invertible transformations enables the approximation of arbitrary complex distributions. RealNVP~\cite{dinh2014nice} introduced affine transformations to boost the representational power of flow models for complex distributions. MAF~\cite{papamakarios2017masked} proposed autoregressive flows to implement sequential variable transformation and ensure invertibility. Glow~\cite{kingma2018glow} leveraged invertible convolutional flows for image generation tasks. Neural ODE~\cite{chen2018neural} treats discrete transformation modeling as a continuous-time process, pushing the number of transformation steps to infinity. Durkan et al.~\cite{durkan2019neural} proposed spline flows, achieving accurate modeling of complex distributions.

\section{Preliminary}
% In this section, we formally describe the baseline reconstruction-based anomaly detection framework and the normalizing flow model, which serves as the theoretical foundation of autoregressive flows.
 
\subsection{Reconstruction-based time-series anomaly detection}
Given a multivariate time-series data $\boldsymbol{x} \in \mathbb{R}^{T \times C}$ after zero-mean normalization, where $T$ denotes the length of discrete time points and $C$ represents the number of variables, our goal is to identify time points containing anomalies. Specifically, we define an embedding layer $f_{\text{embed}}(\cdot)$ that maps $\boldsymbol{x}$ to a high-dimensional embedded representation $\boldsymbol{x}_{\text{emb}} = f_{\text{embed}}(\boldsymbol{x}) \in \mathbb{R}^{T \times D_\mathrm{emb}}$. We then feed $\boldsymbol{x}_{\text{emb}}$ into the encoder $f_{\mathrm{enc}}(\cdot)$, which is composed of a stack of encoding layers, to obtain the latent representation $\boldsymbol{z} \in \mathbb{R}^{T \times D_\mathrm{enc}}$. Finally, a simple projection layer $f_{\text{proj}}(\cdot)$ transforms $\boldsymbol{z}$ into the reconstructed output $\hat{\boldsymbol{x}} \in \mathbb{R}^{T \times C}$.
During training, we minimize the L2 reconstruction error $\mathcal{L}_{\mathrm{rec}} = \|\boldsymbol{x} - \boldsymbol{\hat{x}}\|_2^2$ to learn normal time-series patterns. In the anomaly detection phase, the model fails to reconstruct anomalous data, resulting in larger errors at anomalous time points $T_a=\{t_1, \cdots, t_{a}\}$. The error at time $t \in T_a$ is defined as $\epsilon_t = \frac{1}{C}\sum_{c=1}^{C}\|\boldsymbol{x}_{t,c} - \hat{\boldsymbol{x}}_{t,c}\|_2^2$. The embedding layer and encoder, which typically incorporate attention mechanisms, form the representation layer, while the projection layer and error calculation constitute the judgment layer. Reconstruction error serves as a universal anomaly criterion for reconstruction-based methods.

\subsection{Normalizing Flow}
Suppose we have two random variables $\boldsymbol{x}\in \mathbb{R}^{D}$ and $\boldsymbol{z}\in\mathbb{R}^{D}$, and an invertible, differentiable transformation function $f$ that maps $\boldsymbol{x}$ to $\boldsymbol{z}$. Using the variable substitution rule, we can derive the transformation relationship between their probability distributions as:
\begin{equation}
    p_{\mathrm{X}}(\boldsymbol{x}) = p_{\mathrm{Z}}\bigl(f(\boldsymbol{x})\bigr) \cdot \left| \det\left( \frac{\partial f(\boldsymbol{x})}{\partial \boldsymbol{x}} \right) \right|^{-1},
\end{equation}
where $\det(\cdot)$ denotes the determinant of a matrix.

To model complex target distributions, Normalizing Flow constructs a composite transformation $f_{\text{NF}}$ by cascading $K$ invertible and differentiable transformations $\{f_k\}_{k=1}^K$. This transformation $f_{\text{NF}}$ maps a variable $\boldsymbol{x}$ in the target space to a variable $\boldsymbol{z}$ in the latent space, i.e.,
$$
\boldsymbol{z}=f_{\text{NF}}(\boldsymbol{x}) = f_K \circ f_{K-1} \circ \cdots \circ f_1(\boldsymbol{x}).
$$
The Jacobian determinant of this composite transformation, denoted as $\boldsymbol{J}_{f_{\text{NF}}}(\boldsymbol{x})=\frac{\partial f_{\text{NF}}(\boldsymbol{x})}{\partial \boldsymbol{x}}$, satisfies the multiplicative property as follows:
\begin{equation}
\det\bigl(\boldsymbol{J}_{f_{\text{NF}}}(\boldsymbol{x})\bigr) = \prod_{k=1}^K \det\bigl(\boldsymbol{J}_{f_k}(\boldsymbol{x}_{k-1})\bigr),
\end{equation}
where $\boldsymbol{x}_0 = \boldsymbol{x}$ is the input to the first transformation $f_1$; $\boldsymbol{x}_k = f_k(\boldsymbol{x}_{k-1})$ represents the output of the $k$-th transformation $f_k$; $\boldsymbol{x}_K = f_K(\boldsymbol{x}_{K-1})=\boldsymbol{z}$ is the final output in the latent space; and $\boldsymbol{J}_{f_k}(\boldsymbol{x}_{k-1}) = \frac{\partial f_k(\boldsymbol{x}_{k-1})}{\partial \boldsymbol{x}_{k-1}} \in \mathbb{R}^{D \times D}$ denotes the Jacobian matrix of $f_k$ with respect to its input $\boldsymbol{x}_{k-1}$.

\section{Methodology}
\begin{figure*}[ht]
    \centering
    \includegraphics[width=1\linewidth]{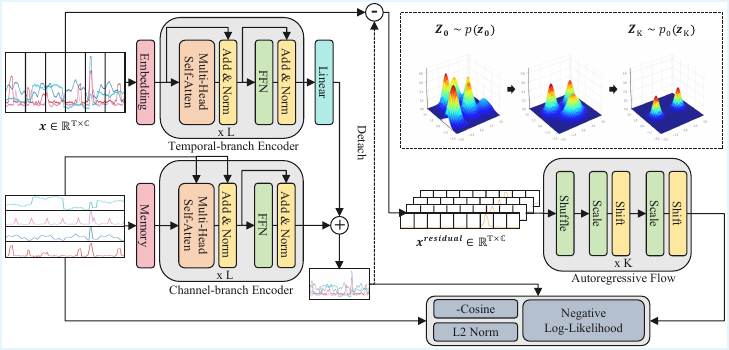}
    \caption{Overall Framework Diagram of Our Method DBR-AF. Without loss of generality, each transformation in our autoregressive flow is represented as a Scale $\&$ Shift operation. The dashed box illustrates the distribution transformation in the autoregressive flow module. Specifically, the reconstructed residual is given by $\boldsymbol{x}^{\mathrm{residual}}=\{\boldsymbol{x}_1^{\mathrm{residual}}, \cdots, \boldsymbol{x}_T^{\mathrm{residual}}\}$, and each residual at time step $t$ is independent and identically distributed (i.i.d.), denoted as $\boldsymbol{z}_0$. The residual $\boldsymbol{z}_0$ is transformed through $K$ layers of autoregressive flow to obtain the latent variable $\boldsymbol{z}_K$, which is assumed to follow the prior distribution $p_0$. Our autoregressive flow model computes the log-likelihood of the original residual $\boldsymbol{z}_0$ by combining the log-likelihood of $\boldsymbol{z}_K$ under the prior $p_0$ and the sum of log-Jacobian determinants accumulated during the invertible transformations.The negative log-likelihood is then used as the optimization objective. The 3D distribution visualization in the dashed box is for illustrative purposes only and does not represent the actual dimensionality of the distribution. In practice, the distribution dimension is typically equal to the number of variables in the temporal signal. 
    }
    \label{fig:4_1}
\end{figure*}

This section introduces our DBR-AF method, with its framework shown in \cref{fig:4_1}. First, we detail the dual branch reconstruction module, which is derived from a basic reconstruction encoder. Next, we present the application of autoregressive flows to anomaly detection for efficient distribution transformation and density estimation. Finally, we summarize the design of the loss function and the implementation of anomaly scoring.
\subsection{Dual-Branch Reconstruction}

For the temporal branch, its structure is based on the representation layer of a basic reconstruction model, with the embedding layer consisting of a 1D convolution and a positional embedding module. The 1D convolution kernel $\boldsymbol{W}_\mathrm{embed} \in \mathbb{R}^{3 \times C \times D_{\mathrm{emb}}}$ maps the channel dimension from $C$ to $D_\mathrm{emb}$ while capturing variations across adjacent time steps. The positional embedding module $\boldsymbol{P}\in \mathbb{R}^{T \times D_{\mathrm{emb}}}$ provides temporal dependency information. The encoder includes $L_{\mathrm{temporal}}$ multi-head self-attention layers, and the projection layer is a fully connected network with a bias term. The temporal branch's reconstruction process is formally expressed as:
$$
\hat{\boldsymbol{x}}^\mathrm{temporal} = f_{\mathrm{enc}}^{\text{temporal}} \left( \boldsymbol{x} \boldsymbol{W}_{\mathrm{embed}} + \boldsymbol{P} \right) \boldsymbol{W}_{\mathrm{proj}} + \boldsymbol{b}_{\mathrm{proj}}.
$$

For the channel branch, we focus on intra-channel statistical characteristics. Since individual channels lack sufficient semantic information, embedding is removed; instead, we transpose the raw input $\boldsymbol{x}$ to obtain $\boldsymbol{x}^{\top} \in \mathbb{R}^{C \times T}$ for subsequent processing. As only a segment of temporal data is used as input for reconstruction, we introduce a memory bank to retain channel-wise features of the full time series. The memory bank $\boldsymbol{\mathcal{M}} \in \mathbb{R}^{M \times T}$, combined with the current transposed segment $\boldsymbol{x}^\top$, is jointly fed into the encoder of the channel branch. The encoder consists of a stack of $L_{\mathrm{channel}}$ cross-attention layers. For the $l$-th layer, we have $\boldsymbol{Q}^{(l)} = \boldsymbol{H}^{(l-1)} \boldsymbol{W}_Q^{(l)}$, $\boldsymbol{K}^{(l)} = \boldsymbol{\mathcal{M}} \boldsymbol{W}_K^{(l)}$, $\boldsymbol{V}^{(l)} = \boldsymbol{\mathcal{M}} \boldsymbol{W}_V^{(l)}$, where $\boldsymbol{W}_{(\cdot)}^{(l)} \in \mathbb{R}^{T \times T}$ denotes a projection matrix. These matrices are then split into $H$ heads for attention computation:
$$
\boldsymbol{Q}_h^{(l)}, \boldsymbol{K}_h^{(l)}, \boldsymbol{V}_h^{(l)} = \text{Split}(\boldsymbol{Q}^{(l)}, \boldsymbol{K}^{(l)}, \boldsymbol{V}^{(l)}, H),\ h\in[1,H],
$$
$$
\text{Attn}_h^{(l)} = \text{Softmax}\left( \frac{\boldsymbol{Q}_h^{(l)} \boldsymbol{K}_h^{(l)\top}}{\sqrt{D_k^{(l)}/H}} \right),
$$
$$
\boldsymbol{H}_{\text{concat}}^{(l)} = \text{Concat}(\text{Attn}_1^{(l)} \boldsymbol{V}_1^{(l)}, ..., \text{Attn}_H^{(l)} \boldsymbol{V}_H^{(l)}).
$$
Finally, the output of the $l$-th layer is defined as:
$$
\boldsymbol{H}^{(l)} = \boldsymbol{H}_{\text{concat}}^{(l)} \boldsymbol{W}_O^{(l)} + \boldsymbol{H}^{(l-1)},
$$
where $\boldsymbol{H}^{(0)} = \boldsymbol{x}^\top$ is the initial input of the encoder, and the final output is $\hat{\boldsymbol{x}}^{\text{channel}} = {\boldsymbol{H}^{(L_{\mathrm{channel}})}}^\top$.
The outputs of the two branches are directly summed to generate the final reconstructed output:
\begin{equation}
    \hat{\boldsymbol{x}} = \hat{\boldsymbol{x}}^{\mathrm{temporal}} + \hat{\boldsymbol{x}}^{\mathrm{channel}}.
\end{equation}

Moreover, since L2 reconstruction loss is prone to being dominated by errors from large-scale variable values, we introduce a cosine similarity distance loss for the full multi-channel data to guide the model to focus more on reconstruction errors in feature orientation:
\begin{equation}
    \mathcal{L}_{\text{rec}} = \|\boldsymbol{x} - \hat{\boldsymbol{x}}\|_2^2 + \lambda \cdot \frac{1}{T} \sum_{t=1}^T \left(1 - \cos\left(\boldsymbol{x}_t, \hat{\boldsymbol{x}}_t\right)\right),
\label{equ: rec}
\end{equation}
where the hyperparameter $\lambda$ adjusts the weight of the two loss components.

\subsection{Autoregressive Flow-based Residual Density Estimation}
Modeling the distribution of low-dimensional encoder outputs and using density estimation for anomaly detection is a widely accepted effective approach\cite{zong2018deep}. However, time-varying characteristics of time series signals (e.g., periodicity and seasonality) manifest as regular fluctuations or distribution shifts over time. In contrast, traditional distribution modeling methods (e.g., Gaussian Mixture Model (GMM) and Kernel Density Estimation (KDE)) rely heavily on the static distribution assumption. These methods presuppose that data follow a fixed global distribution and thus fail to dynamically capture the periodic variations in time series signals. For multivariate scenarios, cross-channel heterogeneity further exacerbates this limitation. Based on this insight, we propose a density estimation strategy for reconstruction residuals.

Specifically, we use the residuals between original inputs and reconstructed outputs as the target for distribution modeling, as these residuals lack the time-varying characteristics of raw time series. Unlike absolute reconstruction errors, these residuals reflect discrepancies between normal and abnormal patterns in both magnitude and direction, defined as:
\begin{equation}
\boldsymbol{x}^{\mathrm{residual}} = \boldsymbol{x} - \mathrm{sg}\left[\hat{\boldsymbol{x}}\right],
\end{equation}
where $\mathrm{sg}\left[\cdot\right]$ denotes the stop-gradient operation~\cite{van2017neural}. This operation prevents the distribution fitting process from interfering with the optimization of time series data reconstruction.

Theoretically, normalizing flows are universal approximators capable of transforming any well-behaved distribution into another through an infinite sequence of invertible transformations~\cite{draxler2024universality}. In this paper, we focus on enhancing the expressive power of these invertible transformations and optimizing the computation of Jacobian determinants to enable efficient and stable model training. To achieve this, we prioritize two practical considerations: first, emphasizing efficient density estimation over the sampling generation process commonly used in generative models, and second, ensuring precise calculation of Jacobian determinants to improve distribution fitting. These considerations guide us toward adopting an autoregressive approach for distribution transformation.

In detail, for the time-invariant residuals $\boldsymbol{x}^{\text{residual}} \in \mathbb{R}^{T \times C}$, we treat each time step as an independent sample. In other words, $\boldsymbol{x}^{\text{residual}}$ is regarded as a collection of $T$ i.i.d. samples, with each residual sample $\boldsymbol{x}_t^{\text{residual}}$ corresponding to an observation of the random variable $X^{\text{residual}} \in \mathbb{R}^C$. We then employ the autoregressive flow, a subclass of normalizing flows, which consists of $K$ stacked invertible transformations, denoted as $f_{\mathrm{AF}} = f_K \circ f_{K-1} \circ \cdots \circ f_1$, where $f_{k}$ comprises $C$ dimension-wise autoregressive transformations $f_{k}^c: \mathbb{R}^c \to \mathbb{R}$. The transformation of the $c$-th channel at the $k$-th layer follows the autoregressive rule:
\begin{equation}
z_{k,c} = f_{k}^c\left( \boldsymbol{z}_{k-1, 1:c-1}, z_{k-1,c} \right),
\end{equation}
where $\boldsymbol{z}_{k} = \left[ z_{k,1}, z_{k,2}, \dots, z_{k,C} \right]^T \in \mathbb{R}^C$ denotes the output vector of the $k$-th layer and $\boldsymbol{z}_0 = \boldsymbol{x}_t^{\text{residual}}$ is the initial input. 

For the $k$-th layer $f_{k}$, its Jacobian matrix $\boldsymbol{J}_{k} \in \mathbb{R}^{C \times C}$ is defined as $\boldsymbol{J}_{k}[i,j] = \frac{\partial z_{k,i}}{\partial z_{k-1,j}}$. Owing to the autoregressive constraint that $z_{k,i}$ depends only on $\boldsymbol{z}_{k-1,1:i}$, $\boldsymbol{J}_{k}$ is strictly lower triangular. The log-determinant of $\boldsymbol{J}_k$ thus simplifies to:
\begin{equation}
    \log \left| \det\boldsymbol{J}_{k} \right| = \sum_{c=1}^C \log \left| \frac{\partial f_{k}^c}{\partial z_{k-1,c}} \right|.
\end{equation}

Notably, the channels of multivariate time series data have no inherent sequential order. We therefore perform random channel permutation for each sample to mitigate the autoregressive constraint’s dependency on fixed dimensional ordering. Let $\boldsymbol{z}_K = f_\mathrm{AF}(\boldsymbol{z}_0)$ denote the final latent variable after $K$ layers of transformation. Unlike simple single Gaussian distributions, we assume $\boldsymbol{z}_K$ follows a predefined Gaussian Mixture Model (GMM) as the prior distribution, defined as:
\begin{equation}
    p_0(\boldsymbol{z}_K) = \sum_{m=1}^M \pi_m \cdot \mathcal{N}\left(\boldsymbol{z}_K \mid \boldsymbol{\mu}_m, \boldsymbol{\Sigma}_m\right),
\end{equation}
where $M$ is the number of Gaussian components, $\pi_m$ is the mixing coefficient satisfying $\sum_{m=1}^M \pi_m = 1$ and $\pi_m \ge 0$, $\boldsymbol{\mu}_m \in \mathbb{R}^C$ is the mean vector, and $\boldsymbol{\Sigma}_m \in \mathbb{R}^{C \times C}$ is the positive definite covariance matrix of the $m$-th Gaussian component. Based on the variable substitution rule of normalizing flows, the log-likelihood of the initial residual input $\boldsymbol{z}_0$ is calculated as: 
\begin{equation}
    \log p(\boldsymbol{z}_0) = \log p_0(\boldsymbol{z}_K) + \sum_{k=1}^K \log \left| \det\boldsymbol{J}_{k} \right|. 
\label{equ:4_3_5}
\end{equation}
Finally, the negative log-likelihood loss (NLL loss) for training the autoregressive flow model is defined as: 
\begin{align}
\mathcal{L}_{\text{nll}} &= -\frac{1}{T}\sum_{t=1}^T \Bigg( \log\Bigg( \sum_{m=1}^M \pi_m \cdot \mathcal{N}\left(\boldsymbol{z}_{K,t} \mid \boldsymbol{\mu}_m, \boldsymbol{\Sigma}_m\right) \Bigg) \notag \\
& \quad + \sum_{k=1}^K \log \left| \det\boldsymbol{J}_{k,t} \right| \Bigg),
\end{align}
where $\boldsymbol{z}_{K,t}$ and $\boldsymbol{J}_{k,t}$ are the final latent variable and Jacobian matrix of the $t$-th residual sample $\boldsymbol{x}_t^{\mathrm{residual}}$, respectively.
Our AF module is implemented based on MAF~\cite{papamakarios2017masked} and adopts the neuron numbering strategy of MADE~\cite{germain2015made} to enable parallel affine transformation of multivariate data while strictly enforcing sequential dependency among variables. Each transformation in the flow is an affine transformation, with the parameters of these affine transformations derived from a two-layer multi-layer perceptron (MLP) network; the hidden layer dimension is set to $D_{\text{AF}}$.

\subsection{Training Loss and Anomaly Evaluation}
Overall, the representation layer of DBR-AF integrates dual-branch reconstruction and autoregressive flow-based distribution transformation of reconstruction residuals. The training loss function is thus defined as:
\begin{equation}
    \mathcal{L} = \mathcal{L}_{\mathrm{rec}} + \beta \mathcal{L}_{\mathrm{nll}},
\end{equation}
where the hyperparameter $\beta$ balances the weights of the NLL loss $\mathcal{L}_{\mathrm{nll}}$ and the reconstruction loss $\mathcal{L}_{\mathrm{rec}}$. For the judgment layer, the anomaly score for each time step $t$ is defined as:
\begin{equation}
    \mathcal{S} = -\log p(\boldsymbol{x}_t - \hat{\boldsymbol{x}}_t) \cdot \Bigg(\|\boldsymbol{x}_t - \hat{\boldsymbol{x}}_t\|_2^2 + \Bigg(1-\cos(\boldsymbol{x}_t, \hat{\boldsymbol{x}}_t)\Bigg)\Bigg),
\end{equation}
where $\log p(\boldsymbol{x}_t - \hat{\boldsymbol{x}}_t)$ is defined in \cref{equ:4_3_5}.

\begin{table*}[htbp]
  \centering
    \caption{Statistics of the seven benchmarks, including the domain of the dataset and the number of feature dimensions, as well as the total length of the training, valid, and test sets. AR denotes the abnormal proportion of the test set.}
  \resizebox{0.7\linewidth}{!}{
  \begin{tabular}{cccccccc}
    \toprule
    Dataset & Domain & Train & Valid & Test & Dimension & AR (\%) \\
    \midrule
    SMD & Server Machine & 566724 & 141681 & 708420 & 38 & 4.2 \\
    MSL & Spacecraft & 46653 & 11664 & 73729 & 55 & 10.5 \\
    SMAP & Spacecraft & 108146 & 27037 & 427617 & 25 & 12.8 \\
    SWaT & Water Treatment & 396000 & 99000 & 449919 & 51 & 12.1 \\
    PSM & Server Machine & 105984 & 26497 & 87841 & 25 & 27.8 \\
    NIPS-TS-GECCO & Water Treatment & 55408 & 13852 & 69261 & 9 & 1.1 \\
    NIPS-TS-Swan & Machinery & 48000 & 12000 & 60000 & 38 & 32.6 \\
    \bottomrule
  \end{tabular}
  }
  \label{tab:dataset_stats}
\end{table*}

\section{Experiments}
\subsection{Setup}
\textbf{Baseline:} We comprehensively benchmark our model against 14 baseline methods, covering classic anomaly detection methods \cite{liu2008isolation, hawkins1980identification}, density estimation methods \cite{breunig2000lof, zong2018deep, yairi2017data}, and reconstruction-based methods. Among these baselines, A.T.~\cite{xu2021anomaly} and DCdetector~\cite{yang2023dcdetector} adopt contrastive learning; TFMAE~\cite{fang2024temporal}, CATCH~\cite{wu2024catch}, and MtsCID~\cite{xie2025multivariate} fuse time-frequency information; MEMTO~\cite{song2023memto} employs a memory mechanism; D3R~\cite{wang2023drift} and EDAD~\cite{zhang2025encode} implement time series decomposition; and DMamba~\cite{chen2024joint} adapts the Mamba framework to MTSAD.

\textbf{Datasets:} We conduct comprehensive evaluations of our method on seven public benchmark datasets. Five of these are well-established real-world benchmarks widely used in the multivariate time series anomaly detection field, namely SMD~\cite{su2019robust}, MSL~\cite{hundman2018detecting}, SMAP~\cite{hundman2018detecting}, SWaT~\cite{mathur2016swat}, and PSM~\cite{abdulaal2021practical}. Complementing these are two additional challenging datasets with more sophisticated anomaly patterns: NIPS-TS-Swan~\cite{angryk2020swan} and NIPS-TS-GECCO~\cite{rehbach2018gecco}. ~\cref{tab:dataset_stats} lists statistics of the 7 public benchmark datasets.

\textbf{Metric:} We adopt threshold-dependent anomaly evaluation metrics with Point Adjust (PA), including Accuracy, Recall, and F1-score, which are widely used in existing literature despite their inherent evaluation biases~\cite{kim2022towards, huet2022local}. Additionally, we utilize AUC-ROC (Area Under the Receiver Operating Characteristic Curve) and AUC-PR (Area Under the Precision-Recall Curve) to facilitate more balanced comparisons across different models.
\begin{itemize}
    \item \textbf{Precision} is the proportion of observations predicted correctly to be abnormal among those predicted to be abnormal, calculated as: $P = \frac{TP}{TP + FP}$.

    \item \textbf{Recall} is the proportion of observations predicted correctly to be abnormal among those labeled to be abnormal, calculated as: $R = \frac{T P}{T P + F N}$.
    \item \textbf{F1-score} can comprehensively evaluate the detecting performance. It is the harmonic mean of precision and recall, calculated as: $F1 = 2 \cdot \frac{P \cdot R}{P + R}$.
    \item \textbf{AUC-ROC} is the area under the ROC curve which depicts the relationship between the false positive rate (FPR) and the true positive rate (TPR) where  
    $TPR = \frac{TP}{T P + F N},$ and $FPR = \frac{F P}{F P + T N}$.
    \item \textbf{AUC-PR} is the area under the Precision-Recall curve which is more suitable to evaluate imbalanced classifiers. 
\end{itemize}

\begin{table*}[h]
\centering
\caption{Performance comparison of different methods on five MTSAD benchmark datasets (P: Precision, R: Recall, F1: F1-score). A dash ($-$) denotes missing performance data for benchmark datasets not evaluated by these methods. \textbf{Bold} entries indicate optimal performance, and \underline{underlined} entries indicate suboptimal performance.}
\label{tab:5_2_1}
\resizebox{\linewidth}{!}{
\begin{tabular}{c |ccc |ccc |ccc |ccc |ccc} 
\hline
\textbf{Dataset} & \multicolumn{3}{c|}{\textbf{SMD}} & \multicolumn{3}{c|}{\textbf{MSL}} & \multicolumn{3}{c|}{\textbf{SMAP}} & \multicolumn{3}{c|}{\textbf{SWaT}} & \multicolumn{3}{c}{\textbf{PSM}} \\
\hline
\textbf{Metric}  & \textbf{P} & \textbf{R} & \textbf{F1} & \textbf{P} & \textbf{R} & \textbf{F1} & \textbf{P} & \textbf{R} & \textbf{F1} & \textbf{P} & \textbf{R} & \textbf{F1} & \textbf{P} & \textbf{R} & \textbf{F1} \\
\hline
Isolation Forest  &42.31 &73.29 &53.64 &53.94 &86.54 &66.45 &52.39 &59.07 &55.53 &49.29 &44.95 &47.02 &76.09 &92.45 &83.48 \\
OC-SVM           &44.34 &76.72 &56.19 &59.78 &86.87 &70.82 &53.85 &59.07 &56.34 &45.39 &49.22 &47.23 &62.75 &80.89 &70.67  \\
LOF              &56.34 &39.86 &46.68 &47.72 &85.25 &61.18 &58.93 &56.33 &57.60 &72.15 &65.43 &68.62 &57.89 &90.49 &70.61   \\
DAGMM           &67.30 &49.89 &57.30 &89.60 &63.93 &74.62 &86.45 &56.73 &68.51 &89.92 &57.84 &70.40 &93.49 &70.03 &80.08 \\
MPPCACD         &71.20 &79.28 &75.02 &81.42 &61.31 &69.95 &88.61 &75.84 &81.73 &82.52 &68.29 &74.73 &76.25 &78.35 &77.29  \\
A.T.             &89.40  &95.45  &92.33  &92.09  &95.15  &93.59  &94.13  &\underline{99.40}  &96.69  &91.55  &96.73  &94.07  &96.91  &98.90  &97.89  \\
MEMTO             &88.24 &\underline{96.16} &92.03 &91.95 &97.23 &94.56 &93.66 &\textbf{99.73} &96.60 &94.28 &91.72 &93.73 &97.47 &98.60 &98.03  \\
DCdetector       &86.08 &85.60 &85.84 &92.09 &\textbf{98.89} &\underline{95.37} &94.42 &98.95 &96.63 &93.29 &\textbf{100.00} &96.53 &97.24 &97.72 &97.48   \\
D3R          &87.74 &96.09 &91.91 &91.77 &94.33 &93.03 &92.23 &96.11 &94.21 &83.09 &83.00 &83.04 &93.84 &\underline{99.11} &96.45  \\
TFMAE            &91.41 &91.07  &91.24  &92.83  &\underline{97.59}  &{95.15}  &94.71  &99.19  &96.90  &\textbf{96.77}  &\textbf{100.00}  &\textbf{98.36}  &98.06  &99.06 &\underline{98.56}  \\
DMamba        &\underline{92.57} &54.04 &68.24 &\textbf{93.69} &64.06 &76.09 &{95.10} &52.98 &68.05 &94.11 &86.75 &90.28 &\textbf{98.66} &82.59 &89.91   \\
MtsCID             &91.50  &95.37  &\underline{93.39}  &93.37  &96.97  &95.13  &\underline{95.90}  &98.79  &\textbf{97.32}  &94.16  &\underline{99.82}  &96.91  &\underline{98.57}  &98.51  &98.54  \\
EDAD              &-  &-  &-  &93.10  &96.10  &94.60  &\textbf{97.00}  &97.40  &\underline{97.20}  &93.80  &\textbf{100.00}  &96.80  &97.80  &98.40  &98.10  \\
CATCH &45.38 &\textbf{98.22} &62.08 &66.86 &94.28 &78.24 &82.42 &56.88 &67.31 &84.94 &90.39 &87.58 &96.68 &98.23 &97.45 \\
% H-PAD             &\underline{92.86}  &\underline{98.20}  &\textbf{95.45}  &\textbf{94.05}  &96.88  &\underline{95.45}  &\underline{96.00}  &98.45  &\underline{97.21}  &94.34  &\textbf{100.00}  &{97.09}  &\textbf{98.82}  &\underline{99.41}  &\textbf{99.12}  \\
Ours       &\textbf{93.77}  &95.85  &\textbf{94.80}  &\underline{93.66}  &\underline{97.59}  &\textbf{95.59}  &93.59  &99.19  &96.31  & \underline{95.43} & \textbf{100.00} & \underline{97.66} &{98.42} &\textbf{99.83}  &\textbf{98.88}  \\
\hline
\end{tabular}
}
\end{table*}

\textbf{Implement Details:}
In this paper, all experiments, parameter sensitivity analyses, and ablation studies of the DBR-AF algorithm were implemented in PyTorch and executed on a single NVIDIA GeForce RTX 4090 GPU with 24 GB of memory. For the optimization of DBR-AF network parameters, the Adam optimizer was employed, with an initial learning rate of $2\mathrm{e}{-2}$ configured for the SMAP dataset, $2\mathrm{e}{-3}$ for the PSM dataset, $2\mathrm{e}{-4}$ for the MSL dataset, and $1\mathrm{e}{-4}$ for other benchmark datasets. The training process of DBR-AF integrated an early stopping mechanism, under which model training is terminated if the reconstruction loss defined in \cref{equ: rec} does not decrease for 5 consecutive epochs.
A non-overlapping sliding window with a length of 100 was utilized to segment the time series data of each dataset. The training dataset was partitioned into a training subset ($80\%$) and a validation subset ($20\%$), with a batch size of 256 specified for model training, meaning that 256 samples are processed per iteration. Consistent key hyperparameters were set across all datasets, including the encoder dimension $D_{\text{enc}}$ of the temporal branch, the number of heads $H$ in the channel-branch encoder, the number of layers $K$ in the AF module, the hidden layer dimension $d_{\text{AF}}$ of the AF module, and the size $M$ of the memory bank. Specifically, the hyperparameter combination $(D_{\text{enc}}, H, K, d_{\text{AF}}, M)$ was fixed at $(512, 4, 2, 64, 100)$ for all experiments, with the exception of ablation studies designed to evaluate the impact of individual hyperparameters.

% For the comparative baseline models, two strategies were employed. First, we reproduced these models exactly according to the parameter settings reported in their original publications. Second, for methods lacking publicly available source code, we directly used the experimental results provided in their official papers.

\subsection{Results}
As shown in \cref{tab:5_2_1}, our proposed method achieves highly competitive performance across all five datasets, with SOTA results on four benchmarks. More evaluation results are presented in \cref{tab:5_2_2,tab:5_2_3}, with our method achieving SOTA performance on five of seven benchmarks. Notably, on the five real-world datasets, our method’s average AUC-ROC and AUC-PR outperform the previous best method by 12.58\% and 10.73\%, respectively.

\begin{table*}[h]
\centering
\caption{Performance comparison of different methods on five MTSAD benchmark datasets (A-R: AUC-ROC, A-P: AUC-PR).}
\label{tab:5_2_2}
\resizebox{0.8\linewidth}{!}{
\begin{tabular}{c |cc |cc |cc |cc |cc | cc} 
\hline
\textbf{Dataset} & \multicolumn{2}{c|}{\textbf{SMD}} & \multicolumn{2}{c|}{\textbf{MSL}} & \multicolumn{2}{c|}{\textbf{SMAP}} & \multicolumn{2}{c|}{\textbf{SWaT}} & \multicolumn{2}{c|}{\textbf{PSM}} & \multicolumn{2}{c}{\textbf{Avg}}\\
\hline
\textbf{Metric}  & \textbf{A-R} & \textbf{A-P} & \textbf{A-R} & \textbf{A-P} & \textbf{A-R} & \textbf{A-P} & \textbf{A-R} & \textbf{A-P} & \textbf{A-R} & \textbf{A-P} &\textbf{A-R} &\textbf{A-P}\\
\hline
A.T.             &47.28 &3.70 &48.72 &10.64 &49.67 &12.50 &29.40 &8.82 &48.56 &29.42   &44.73 &13.02   \\
MEMTO           &  73.24 & 10.35 & 49.99 & 10.48 & 59.59 & \underline{16.29} & 45.41 & 11.45 & 49.75&  26.96 & 55.60 & 15.11 \\
DCdetector       &48.77 &{4.19} &50.06 &10.61 &48.87 &12.48 &49.74 &11.60 &49.83 &27.64  &49.45  &13.30  \\
TFMAE &35.76 &3.68 &51.93 &11.06 &54.96 &14.56 &50.22 &12.13 &48.21 &27.28 &48.22 &13.74 \\
D3R              &64.20 &12.24 &\underline{65.26} &\underline{16.99} &41.35 &10.62 &56.65 &13.30 &50.03 &26.31 &55.50 &15.89 \\
DMamba     &64.55 &11.99 &61.54 &15.02 &38.99 &10.85 &\underline{74.49} &\underline{25.33} &59.53 &40.17  &59.82 &20.67\\
CATCH           &\underline{78.74}  &\underline{16.35}  &\textbf{66.21}  &14.53  &53.11  &14.37  &39.19  &22.66  &65.83  &43.96  &60.62  &22.37  \\
EDAD          &-  &-  &61.90  &\textbf{19.70}  &\underline{59.90} &14.70  &57.10 &17.20  &\underline{66.90}  &\underline{51.70}  &\underline{61.45}  &\underline{25.83}  \\
% H-PAD             &76.49  &14.05  &59.99  &15.06  &59.13  &15.30  &\underline{81.54}  &\underline{53.99}  &\underline{75.01}  &\underline{51.83}  & \underline{70.43} & \underline{30.05} \\
Ours        &\textbf{81.59}  &\textbf{19.58}  & 59.67 &15.40  &\textbf{61.83}  &\textbf{16.60} &\textbf{82.29}  
&\textbf{72.36}  &\textbf{79.81}  &\textbf{58.57}  &\textbf{73.03}  &\textbf{36.50}  \\
\hline
\end{tabular}
}
\end{table*}

\begin{table}[ht]
\centering
\caption{Performance comparison of different methods on the other two MTSAD benchmark datasets.}
\label{tab:5_2_3}
\resizebox{1\linewidth}{!}{
\begin{tabular}{c |ccc |ccc } 
\hline
\textbf{Dataset} & \multicolumn{3}{c|}{\textbf{NIPS-TS-Swan}} & \multicolumn{3}{c}{\textbf{NIPS-TS-GECCO}}\\
\hline
\textbf{Metric}   & \textbf{A-R}  & \textbf{A-P} & \textbf{F1}& \textbf{A-R}  & \textbf{A-P} &\textbf{F1}\\
\hline
A.T.              &43.49 &28.62 &73.09 &33.46 &1.48  &19.20\\
MEMTO            &51.12 &49.06 &73.91 &60.96 &4.21 &{69.60}\\
DCdetector        &48.50 &31.71 &73.56 &50.53 &1.72 &28.35\\
TFMAE  &48.35 &32.18 &73.43 &49.29 &1.07 &42.87\\
D3R                &53.40 &40.97 &67.57 &80.32 &12.39 &58.66\\
DMamba       &\underline{77.84} &\underline{64.64} &74.53 &\underline{96.93} &\underline{46.32} &44.23\\
EDAD             &50.10 &32.60 &73.90 &- &- &-\\
CATCH          &52.03 &44.46 &\textbf{75.85} &\textbf{97.22} &\textbf{49.20} &\textbf{79.48}\\
% H-PAD             &\underline{81.66} &\underline{74.31} &- &75.96 &7.30 &-\\
Ours         &\textbf{84.20} &\textbf{77.92} &\underline{74.87}  &86.59 &33.72 & \underline{76.91}\\
\hline
\end{tabular}
}
\end{table}

\begin{table}[htbp]
\centering
\caption{"Channel-only" denotes the configuration retaining only the channel-wise reconstruction branch; "cosine" refers to the cosine similarity constraint; "residual" indicates that the input reconstruction residual is adopted as the input of AF; "detach" represents the disconnection of gradient propagation from AF to DBR; and "shuffle" corresponds to random channel permutation.}
\label{tab:5_3_2}
\resizebox{1\linewidth}{!}{
\begin{tabular}{ccccc |ccc } 
\hline
\multicolumn{5}{c|}{\textbf{Component}} & \multicolumn{3}{c}{\textbf{Avg}} \\
\hline
 channel-only &cosine &residual &detach &shuffle & \textbf{A-R} & \textbf{A-P} & \textbf{F1} \\
\hline

\ding{55} &\ding{51} &\ding{51} &\ding{51} &\ding{51} &\textbf{73.03} &\textbf{36.50} &\textbf{96.65}\\

\ding{51} &\ding{51} &\ding{51} &\ding{51} &\ding{51} &61.60\tiny{(\textcolor{red}{-11.43})} &28.82\tiny{(\textcolor{red}{-7.68})} &{94.82}\tiny{(\textcolor{red}{-1.83})}\\

\ding{55} &\ding{55} &\ding{51} &\ding{51} &\ding{51} &70.05\tiny{(\textcolor{red}{-2.98})} &35.14\tiny{(\textcolor{red}{-1.36})} &93.85\tiny{(\textcolor{red}{-2.80})}\\

\ding{55} &\ding{51} &\ding{55} &\ding{51} &\ding{51} &70.19\tiny{(\textcolor{red}{-2.84})} &34.34\tiny{(\textcolor{red}{-2.16})} &96.16\tiny{(\textcolor{red}{-0.49})}\\

\ding{55} &\ding{51} &\ding{51} &\ding{55} &\ding{51} &66.76\tiny{(\textcolor{red}{-6.27})} &32.61\tiny{(\textcolor{red}{-3.89})} &96.06\tiny{(\textcolor{red}{-0.59})}\\

\ding{55} &\ding{51} &\ding{51} &\ding{51} &\ding{55} &71.66\tiny{(\textcolor{red}{-1.37})} &34.98\tiny{(\textcolor{red}{-1.52})} &96.31\tiny{(\textcolor{red}{-0.34})}\\

\hline
\end{tabular}
}
\end{table}

\begin{table*}[htbp]
\centering
\caption{DBR denotes the dual branch reconstruction encoder, and AF the autoregressive flow model.}
\label{tab:5_3_1}
\resizebox{1\linewidth}{!}{
\begin{tabular}{cc |ccc |ccc |ccc |ccc |ccc | ccc} 
\hline
\multicolumn{2}{c|}{\textbf{Module}} & \multicolumn{3}{c|}{\textbf{SMD}} & \multicolumn{3}{c|}{\textbf{MSL}} & \multicolumn{3}{c|}{\textbf{SMAP}} & \multicolumn{3}{c|}{\textbf{SWaT}} & \multicolumn{3}{c|}{\textbf{PSM}} & \multicolumn{3}{c}{\textbf{Avg}}\\
\hline
 DBR &AF  & \textbf{A-R} & \textbf{A-P} & \textbf{F1} & \textbf{A-R} & \textbf{A-P} & \textbf{F1} & \textbf{A-R} & \textbf{A-P} & \textbf{F1} & \textbf{A-R} &\textbf{A-P} &\textbf{F1} &\textbf{A-R} &\textbf{A-P} &\textbf{F1} &\textbf{A-R} &\textbf{A-P} &\textbf{F1}\\
\hline
 \ding{55}& \ding{55} &77.19 &17.27 &88.16  &\textbf{62.46}  &\textbf{15.60} &88.84 &39.74  &10.11 &71.72  &81.89  &\textbf{72.71}  &87.66  &75.41  &53.85 &97.83 &67.34 &33.92 &86.84\\

 \ding{51}&\ding{55} &80.54 &18.66 &88.89  &51.87  &13.21 &92.14 &\textbf{61.98} &\textbf{16.89} &\underline{96.16}  &81.94  &72.21  &89.43  &60.45  &34.40 &97.36 &67.36 &31.07 &92.80\\
 
 \ding{55}&\ding{51} &\underline{80.87} &\underline{19.39} &\underline{94.78}  &\underline{61.82}  &15.39 &\underline{93.26} &53.13 &12.59 &86.10  &\underline{82.05}  &\underline{72.69}  &\underline{97.25}  &\underline{78.70}  &\underline{57.84} &\textbf{98.95} &\underline{71.31} &\underline{35.58} &\underline{94.07}\\

\ding{51} &\ding{51} &\textbf{81.59} &\textbf{19.58} &\textbf{94.80}  &59.67  &\underline{15.40} &\textbf{95.59} &\underline{61.83}  &\underline{16.60} &\textbf{96.31}  &\textbf{82.29}  &72.36  &\textbf{97.66}  &\textbf{79.81}  &\textbf{58.57} &\underline{98.88} &\textbf{73.03} &\textbf{36.50} &\textbf{96.65}\\
\hline
\end{tabular}
}
\end{table*}

We conduct ablation experiments on five real-world datasets to systematically evaluate the contribution of each core module, with detailed results shown in \cref{tab:5_3_1}. Experimental results indicate that the baseline reconstruction model, which relies solely on the temporal encoder, achieves moderate performance but yields a notably low F1-score. Incorporating the dual-branch reconstruction mechanism significantly enhances the model's precision and recall for anomaly detection, improving the F1-score by 6.04\%. Furthermore, integrating autoregressive normalizing flow-based density estimation brings about a comprehensive performance boost. Specifically, the AUC-ROC and AUC-PR metrics exhibit strong discriminative capability in distinguishing normal from abnormal instances. The synergistic combination of these two modules enables our method to achieve highly competitive performance across multiple metrics on diverse datasets, with the AUC-ROC, AUC-PR, and F1-score improved by 5.69\%, 2.58\%, and 9.81\%, respectively.

In addition to the aforementioned experiments, we conduct ablation of individual components across different modules to quantitatively evaluate their respective impacts on overall model performance. Retaining only the channel branch reconstruction significantly impairs the model’s ability to capture time-varying patterns and multivariate relationships, reducing its discriminative power between normal and abnormal samples. As shown in \cref{tab:5_3_2}, the AUC-ROC and AUC-PR decrease by 11.43\% and 7.68\%, respectively. Notably, the F1-score only drops by 1.83\%, indicating the model still assigns higher anomaly scores to abnormal samples. Ablating the cosine similarity constraint leads to the largest degradation in F1-score (2.8\% reduction). This phenomenon occurs because the model can no longer effectively detect subtle low-scale anomalies without this constraint. Density estimation on residuals enables more effective discrimination between normal and abnormal samples. Allowing gradients to flow from the AF module to the DBR module causes the model to prioritize distribution alignment over reconstruction, which underscores that reconstruction is a more fundamental prerequisite for the model to capture normal patterns. Removing the shuffle operation causes the model to memorize the fixed channel order during training, impairing its generalization ability.

\section{Further Analysis}
\subsection{Discussion about the Autoregressive Flow Module}
\subsubsection{Dimension of hidden layers in the Affine Transformation Network}
\begin{figure}[ht]
    \centering
    \begin{tabular}{cc}
        \begin{minipage}{0.23\textwidth}
            \centering
            \includegraphics[width=\textwidth]{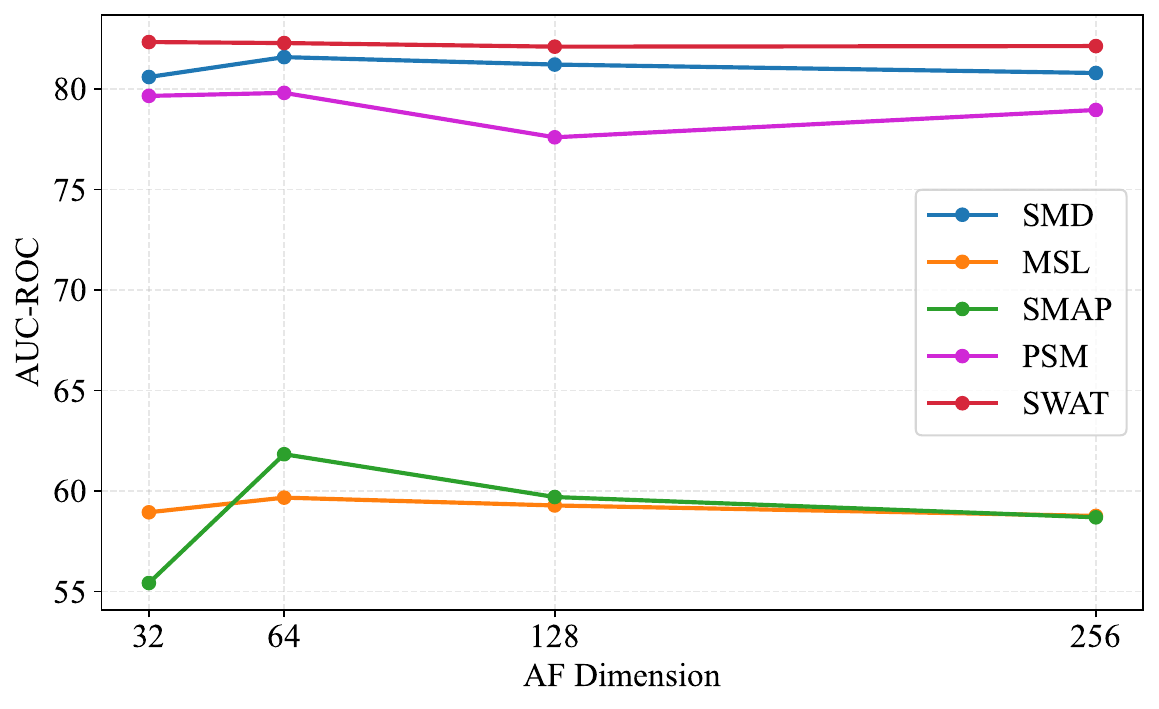}
            % \subcaption{}
        \end{minipage} 
        \hfill
        \begin{minipage}{0.23\textwidth}
            \centering
            \includegraphics[width=\textwidth]{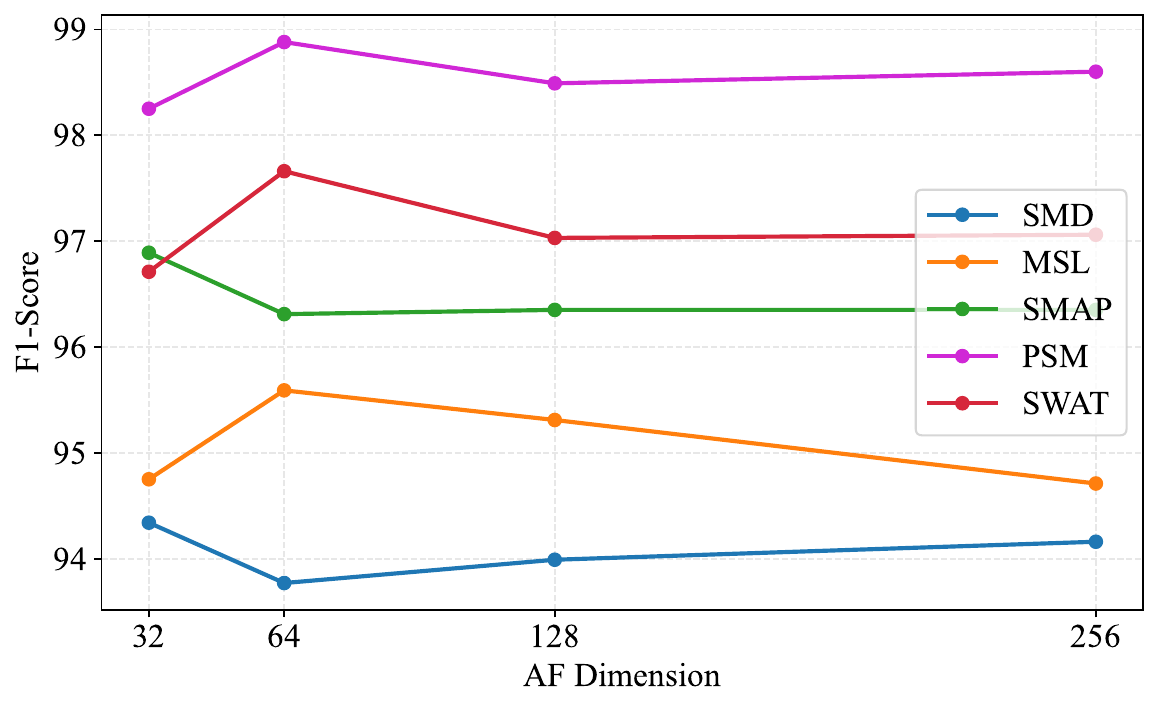}
            % \subcaption{}
        \end{minipage} 
    \end{tabular}
    \caption{Variations of F1-score with $D_{\mathrm{AF}}$.}
    \label{fig:6_4}
\end{figure}

The MLP network for affine transformation in the AF module generates the corresponding scale and shift parameters based on the input. The dimension \(D_{\text{AF}}\) of its hidden layer affects the distribution transformation and fitting performance. We evaluated the model performance under different hidden layer dimensions of \(D_{\text{AF}}\) to quantify this impact. As shown in \cref{fig:6_4}, multivariate time series datasets differ in channel count, resulting in varying autoregressive flow transformation performance across different hidden layer dimensions. In our method, we uniformly set the hidden layer dimension to 64, which achieves balanced performance across all datasets.

\subsubsection{Different Prior distributionint in the Autoregressive Flow Module}
For the prior distribution used in the AF module, we introduce two candidate distributions for comparative performance evaluation: the diagonal Gaussian distribution and the Gaussian mixture distribution with a variable number of components. Additionally, each of these prior distributions is configured in either a frozen state (with fixed parameters) or a trainable state (with parameters optimized during model training). Coefficients for the Gaussian mixture distribution are set to a constant value of $\frac{1}{n_\mathrm{c}}$. Detailed performance comparison results are presented in \cref{tab:prior_performance}. Experimental results show that the choice of prior distribution has a limited impact on overall model performance. Notably, the two-component Gaussian mixture distribution in a frozen state achieves the best average performance across multiple datasets. This finding indicates that a more complex prior distribution does not necessarily improve model performance, which may be attributed to overfitting during training when the prior distribution is overly intricate.

\begin{table}[htbp]
  \centering
      \caption{Performance comparison of the AF module with different prior distributions in our method.}
  \resizebox{1\linewidth}{!}{
  \begin{tabular}{c c c c c c c}
    \toprule
    Prior Distribution &Trainable & MSL & SMAP & PSM & SWaT & SMD \\
    \midrule
    \multirow{2}{*}{Diagonal Gaussian} & \ding{55}& 95.20 & 94.70 & 97.11 & 97.23 & 94.03 \\
    & \ding{51}& 95.25 & 94.55 & 97.02 & 97.40 & 94.18 \\
    \midrule
    \multirow{2}{*}{Gaussian Mixture ($n_\mathrm{c}$ = 2)} & \ding{55}& 95.59 & \textbf{96.31} & \textbf{98.88} & 97.66 & 94.80 \\
    & \ding{51}& 95.26 &95.74 &98.80 & 97.83 & 94.37 \\
    \midrule
    \multirow{2}{*}{Gaussian Mixture ($n_\mathrm{c}$ = 3)} & \ding{55}& \textbf{96.02} & 94.93 & 97.44 & 97.79 & \textbf{95.01} \\
    & \ding{51} & 95.42 & 95.21 & 98.12 & 97.83 & 94.91 \\
    \midrule
    \multirow{2}{*}{Gaussian Mixture ($n_\mathrm{c}$ = 4)} & \ding{55}& 95.85 & 94.21 & 96.72 & 97.89 & 94.85 \\
    & \ding{51}& 95.09 & 94.42 & 96.99 & \textbf{97.90} & 94.88 \\
    \bottomrule
  \end{tabular}
}
  \label{tab:prior_performance}
\end{table}

\subsubsection{Compare with Gaussian Mixture Model}
We also conducted additional experiments to comprehensively compare the AF module with traditional density estimation methods such as the Gaussian Mixture Model (GMM). Notably, instead of optimizing GMM using the Expectation-Maximization (EM) algorithm, we directly replaced the AF module with GMM for end-to-end optimization, and the settings of relevant hyperparameters for GMM were kept consistent with those of the AF module. Performance comparison results are presented in \cref{tab:AF_GMM_performance}. Experimental results show that the AF module consistently outperforms GMM across multiple datasets in both F1 score and AUC ROC metrics. This finding highlights the superior ability of autoregressive normalizing flows to capture complex data distributions compared to traditional density estimation methods.

\begin{table}[ht]
  \centering
  \caption{Performance comparison between GMM (direct replacement) and the AF module.}
  \resizebox{1\linewidth}{!}{
  \begin{tabular}{c cc cc cc cc cc}
    \toprule
    \multirow{2}{*}{\textbf{Method}} & \multicolumn{2}{c}{\textbf{MSL}} & \multicolumn{2}{c}{\textbf{SMAP}} & \multicolumn{2}{c}{\textbf{PSM}} & \multicolumn{2}{c}{\textbf{SWaT}} & \multicolumn{2}{c}{\textbf{SMD}} \\
    \cmidrule(lr){2-3} \cmidrule(lr){4-5} \cmidrule(lr){6-7} \cmidrule(lr){8-9} \cmidrule(lr){10-11}
    & \textbf{F1} & \textbf{A-R} & \textbf{F1} & \textbf{A-R} & \textbf{F1} & \textbf{A-R} & \textbf{F1} & \textbf{A-R} & \textbf{F1} & \textbf{A-R} \\
    \midrule  
    AF & \textbf{95.59} &\textbf{59.67} &96.31 &\textbf{61.83}& \textbf{98.88} &\textbf{79.91} & \textbf{97.66} &\textbf{82.29}& \textbf{94.80} &\textbf{81.59} \\
    GMM & 95.05 &57.67 & \textbf{96.44} &56.29 & 97.66 &78.91 & 96.51 &82.10 & 93.66 &81.15\\
    \bottomrule
  \end{tabular}
  }
  \label{tab:AF_GMM_performance}
\end{table}

\subsubsection{Compare with other Normalizing Flows}
To investigate the impact of different normalizing flows on the model, we conduct experiments to compare the performance of DBR-AF integrated with autoregressive flows and other representative normalizing flow architectures. The specific flow models selected for comparison are detailed as follows:
\begin{itemize}
    \item \textbf{MAF}~\cite{papamakarios2017masked} is a representative autoregressive normalizing flow that uses masked weight matrices to enforce an autoregressive dependency among variables. It enables efficient log Jacobian determinant calculation and excels at density estimation, though its sequential sampling mechanism leads to slower generation speed.
    \item \textbf{NeuralSpline}~\cite{durkan2019neural} constructs invertible transformations via rational quadratic splines, offering strong flexibility to model complex multimodal distributions without relying on affine operations. It is compatible with both autoregressive and coupling-based frameworks, balancing expressive power and computational feasibility across tasks.
    \item \textbf{NICE}~\cite{dinh2014nice} is a pioneering coupling-based flow that only applies additive transformations to split variable groups, simplifying Jacobian calculation to a unit determinant. Its lightweight structure ensures easy implementation, but its limited expressive capacity restricts performance on high-dimensional complex data.
    \item \textbf{RealNVP}~\cite{dinh2016density} improves upon NICE by combining additive and affine transformations, enabling both volume-preserving and non-volume-preserving mappings for flexible data density estimation. It adopts checkerboard/channel-wise masking to handle high-dimensional data like images, achieving a favorable trade-off between model capacity and parallel computation efficiency.
\end{itemize}

Detailed performance comparison results are presented in \cref{tab:NF_performance}. For consistency, the hidden layer dimension is uniformly set to 64. MAF achieves the best performance across 6 out of 10 metrics, as it adopts dimension-wise conditional density estimation for approximation—an approach also used in our work. Neural Spline delivers the second-best overall performance, leveraging its universal density approximation capability enabled by strictly monotonic neural networks. NICE and RealNVP show comparable performance, and their shorter iteration time highlights more efficient density estimation.

\begin{table*}[htbp]
  \centering
      \caption{Performance comparison of DBR-AF integrated with different normalizing flow architectures. Iteration Time denotes the average per-iteration time calculated across five datasets.}
  \resizebox{0.85\linewidth}{!}{
  \begin{tabular}{c c cc cc cc cc cc}
    \toprule
    \multirow{2}{*}{\textbf{Flow Type}} & \multirow{2}{*}{\textbf{Iteration Time(s/it)}} & \multicolumn{2}{c}{\textbf{MSL}} & \multicolumn{2}{c}{\textbf{SMAP}} & \multicolumn{2}{c}{\textbf{PSM}} & \multicolumn{2}{c}{\textbf{SWaT}} & \multicolumn{2}{c}{\textbf{SMD}} \\
    \cmidrule{3-4} \cmidrule(lr){5-6} \cmidrule(lr){7-8} \cmidrule(lr){9-10} \cmidrule(lr){11-12}
       & & \textbf{F1} & \textbf{A-R} & \textbf{F1} & \textbf{A-R}& \textbf{F1} & \textbf{A-R}& \textbf{F1} & \textbf{A-R}& \textbf{F1} & \textbf{A-R} \\
    \midrule
    MAF & 0.080 & \textbf{95.59} &\textbf{59.67} &96.31 &\textbf{61.83}& \textbf{98.88} &\textbf{79.91} & \textbf{97.66} &82.29& 94.80 &81.59 \\
    NerualSpline &0.068 &94.78 &59.20 &\textbf{96.47} &59.21 &98.19 &78.99 &96.63 &82.38 &\textbf{95.15} &\textbf{81.77} \\
    NICE & \textbf{0.052} &95.26 &58.99 &96.22 &58.24 &98.30 &79.58 & 96.56 & \textbf{82.47} &93.77 &79.95 \\
    RealNVP & \textbf{0.052}& 95.30 &59.05& 96.12 &58.42 & 97.60 &78.51 & 95.88 &82.40 & 94.09 &81.10\\
    \bottomrule
  \end{tabular}
  }
  \label{tab:NF_performance}
\end{table*}
\subsection{Sensitivity to key hyperparameters}

\begin{figure}[htp]
    \centering
    \begin{tabular}{cc}
        \begin{minipage}{0.23\textwidth}
            \centering
            \includegraphics[width=\textwidth]{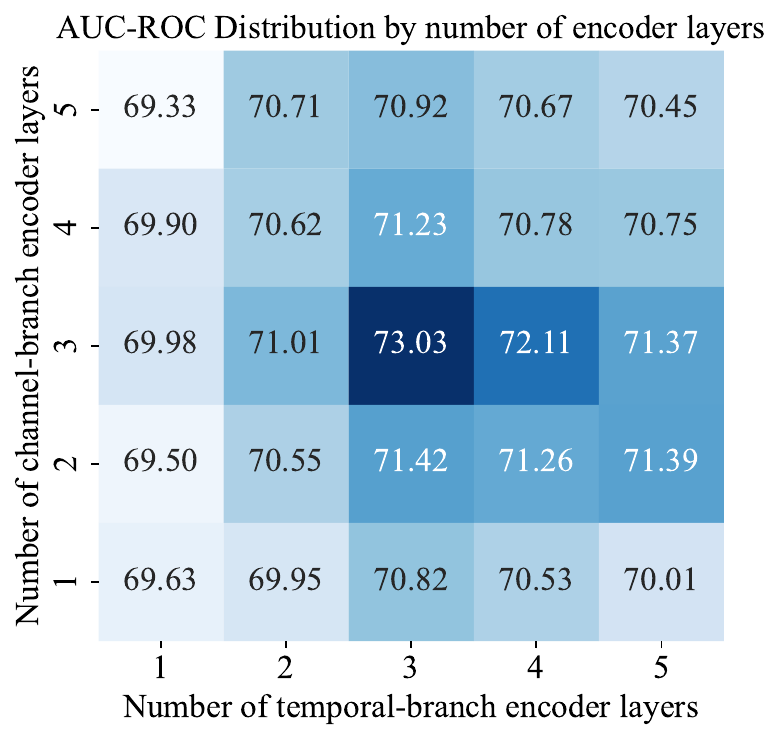}
            % \subcaption{}
        \end{minipage} 
        \hfill
        \begin{minipage}{0.23\textwidth}
            \centering
            \includegraphics[width=\textwidth]{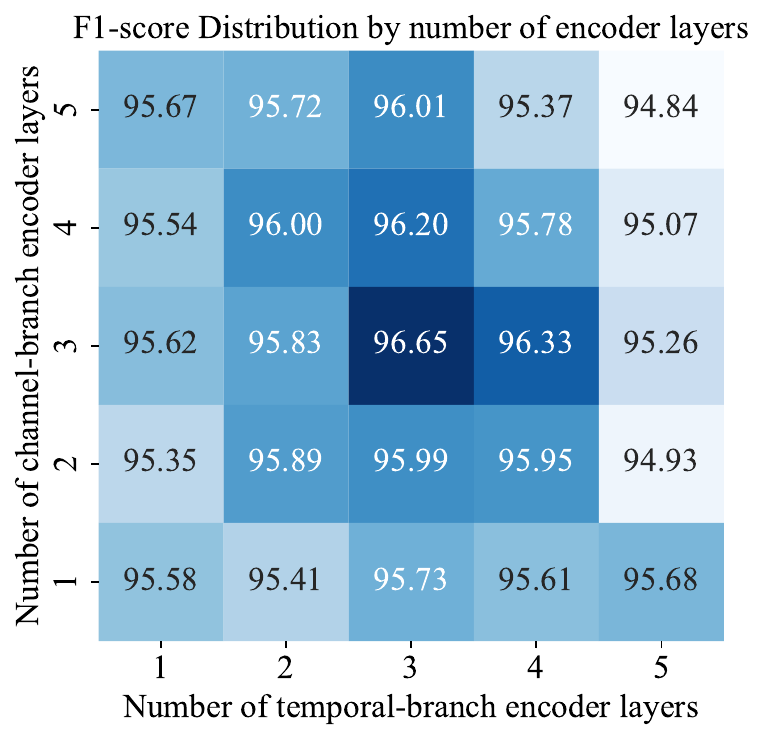}
            % \subcaption{}
        \end{minipage} 
    \end{tabular}
    \caption{AUC-ROC and F1-score distributions by number of encoder Layers, averaged over five real-world benchmark datasets.}
    \label{fig:6_1}
\end{figure}

\subsubsection{Number of Encoder Layers}
The left subfigure of \cref{fig:6_1} illustrates how the model's performance varies with the number of encoder layers in the dual-branch encoder. Increasing the layer count in the temporal branch enhances the model’s ability to differentiate between normal and abnormal samples. In contrast, the channel branch, which focuses on signal reconstruction without high-dimensional transformations, experiences overfitting when the layer count is too large. For the F1-score metric, the impact of different encoder layer configurations on performance is relatively minor, with optimal results achieved when both branches are set to 3 layers. 

\subsubsection{Window Size and Encoder Dimension of Temporal Branch}
\begin{figure}[htbp]
    \centering
    \begin{tabular}{cc}
        \begin{minipage}{0.23\textwidth}
            \centering
            \includegraphics[width=\textwidth]{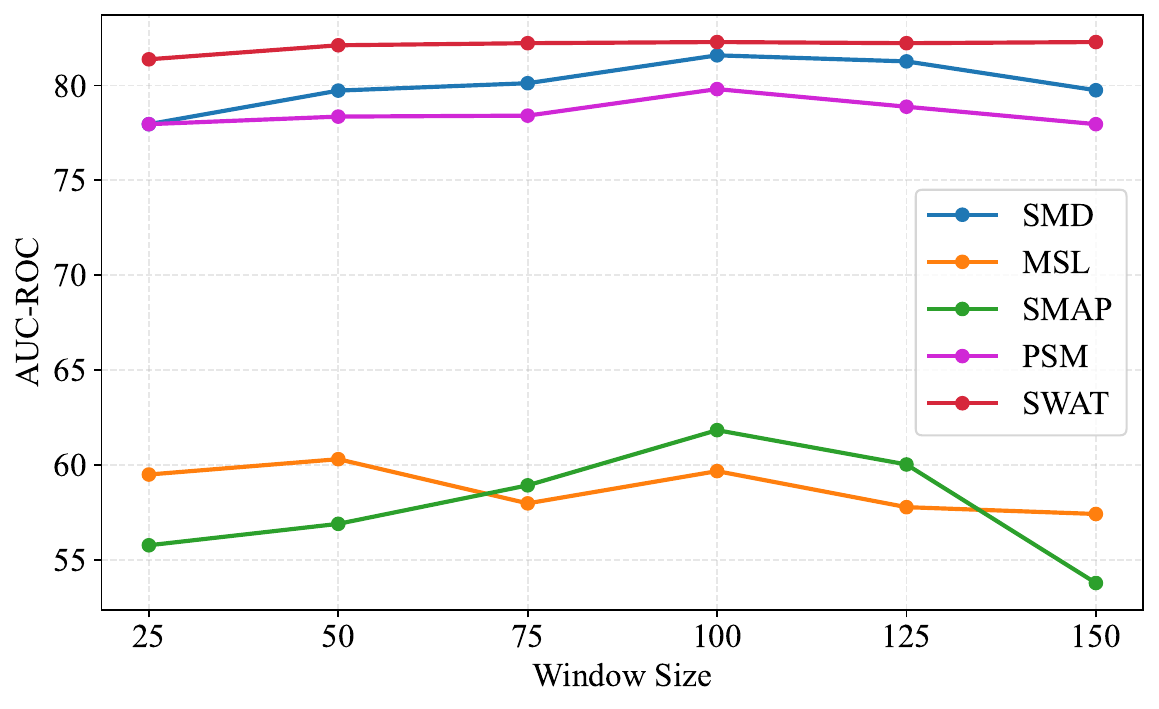}
            % \subcaption{}
        \end{minipage} 
        \hfill
        \begin{minipage}{0.23\textwidth}
            \centering
            \includegraphics[width=\textwidth]{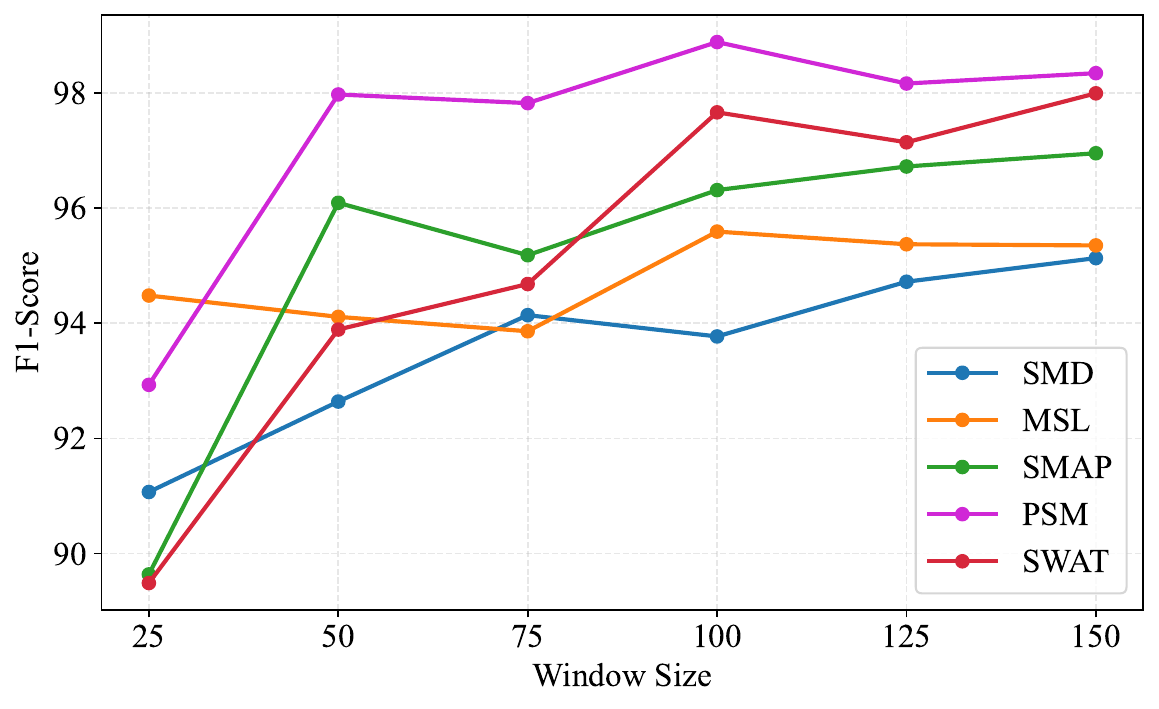}
            % \subcaption{}
        \end{minipage} 
    \end{tabular}
    \caption{Variations of AUC-ROC and F1-score with Window Size.}
    \label{fig:6_2}
\end{figure}

\begin{figure}[htbp]
    \centering
    \begin{tabular}{cc}
        \begin{minipage}{0.23\textwidth}
            \centering
            \includegraphics[width=\textwidth]{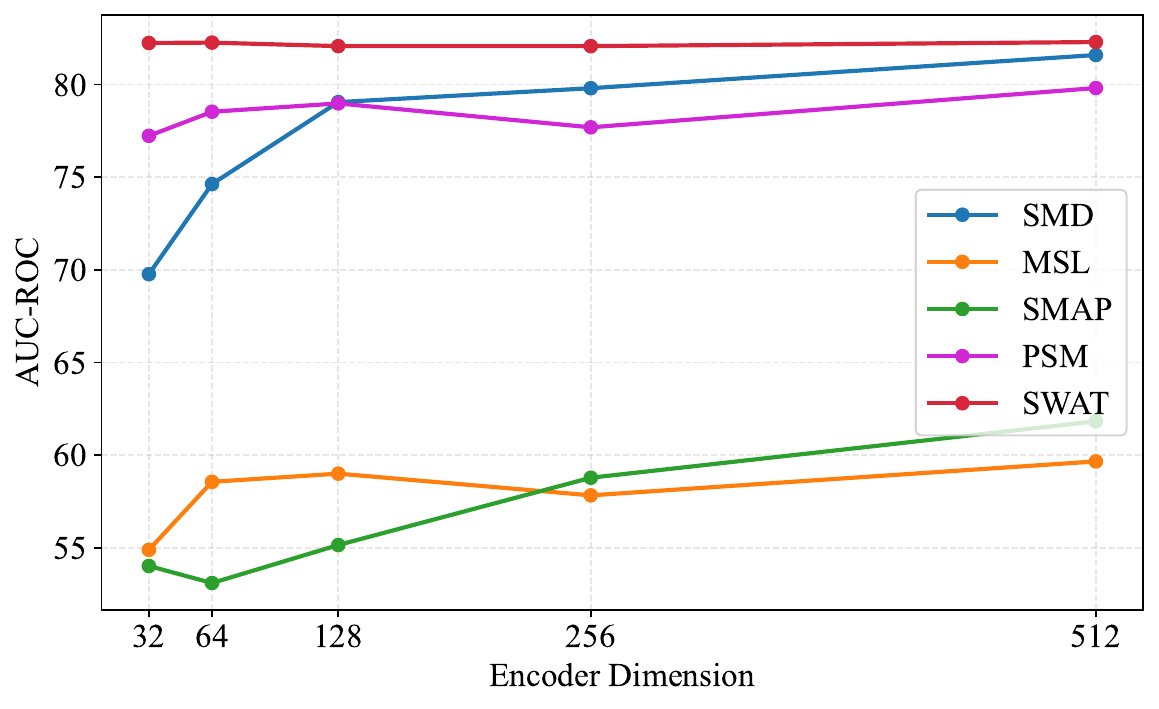}
            % \subcaption{}
        \end{minipage} 
        \hfill
        \begin{minipage}{0.23\textwidth}
            \centering
            \includegraphics[width=\textwidth]{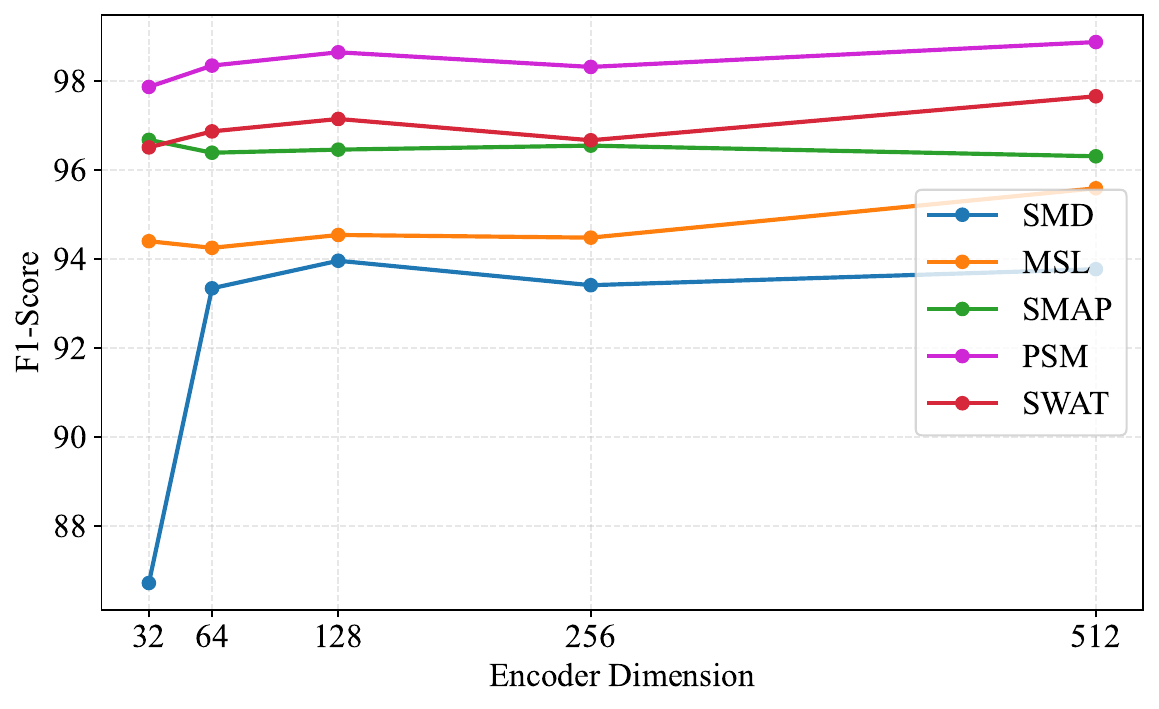}
            % \subcaption{}
        \end{minipage} 
    \end{tabular}
    \caption{Variations of AUC-ROC and F1-score with Encoder Dimension of Temporal Branch $D_{\mathrm{enc}}$.}
    \label{fig:6_3}
\end{figure}

\cref{fig:6_2,fig:6_3} characterize the model's sensitivity to two key hyperparameters, namely window size and temporal branch encoder dimension, across datasets. The window size acts as both the patch length of time series data and the channel branch encoder dimension, making the model more sensitive to its variation than to that of the temporal branch encoder dimension, with optimal performance achieved at a window size of 100. The temporal branch encoder dimension has a limited effect on the model, yet higher dimensions generally better capture time-varying information and multivariate correlations. 

\begin{table}[htbp]
  \centering
  % 左表格
  \begin{minipage}{0.48\linewidth}
    \centering
    \caption{F1-score under Different $\lambda$ Values}
    \resizebox{1\linewidth}{!}{
    \begin{tabular}{cccccc}
    \toprule
    $\lambda$ & MSL & SMAP & PSM & SWaT & SMD \\
    \midrule
    0.4 & 94.14 & 95.98 & 97.75 & 96.34 & 93.10 \\
    0.6 & 93.85 & 96.38 & 98.12 & 96.89 & 93.98 \\
    0.8 & 94.42 & \textbf{96.57} & 98.77 & \textbf{97.68} & 94.02 \\
    1.0 & \textbf{95.59} & 96.31 & \textbf{98.88} & 97.66 & \textbf{94.80} \\
    1.2 & 94.87 & 95.76 & 98.45 & 97.21 & 93.75 \\
    1.4 & 93.92 & 95.42 & 98.03 & 96.75 & 92.90 \\
    \bottomrule
    \end{tabular}
    }
    \label{tab:lambda_performance}
  \end{minipage}
  \hfill
  % 右表格
  \begin{minipage}{0.48\linewidth}
    \centering
    \caption{F1-score under Different $\beta$ Values}
    \resizebox{1\linewidth}{!}{
    \begin{tabular}{cccccc}
    \toprule
    $\beta$ & MSL & SMAP & PSM & SWaT & SMD \\
    \midrule
    0.4 & 92.34 & 95.84 & 97.95 & 97.49 & 91.89 \\
    0.6 & 94.02 & 96.00 & 98.32 & 97.80 & 93.05 \\
    0.8 & 95.12 & \textbf{96.54} & 98.67 & 97.29 & 94.16 \\
    1.0 & \textbf{95.59} & 96.31 & \textbf{98.88} & 97.66 & \textbf{94.80} \\
    1.2 & 94.88 & 95.90 & 98.48 & \textbf{97.85} & 93.72 \\
    1.4 & 93.76 & 95.45 & 98.12 & 96.88 & 92.85 \\
    \bottomrule
    \end{tabular}
    }
    \label{tab:beta_performance}
  \end{minipage}
\end{table}

\subsubsection{Hyperparameter $\lambda$ and $\beta$}
We evaluate model performance across different values of hyperparameters $\lambda$ and $\beta$ for the balanced loss function, with results presented in \cref{tab:lambda_performance} and \cref{tab:beta_performance}. The model demonstrates consistent sensitivity to both $\lambda$ and $\beta$ across multiple datasets, with the MSL and SMD datasets showing the most significant performance variations.

\begin{figure*}[htbp]
    \centering
    % 缩小列间距，让4列能放下且不拥挤
    \setlength{\tabcolsep}{6pt}
    % 3行4列，统一宽度，总宽度约占 A4 一半高度
    \begin{tabular}{cccc}
        % 第 1 行
        \includegraphics[width=0.22\textwidth]{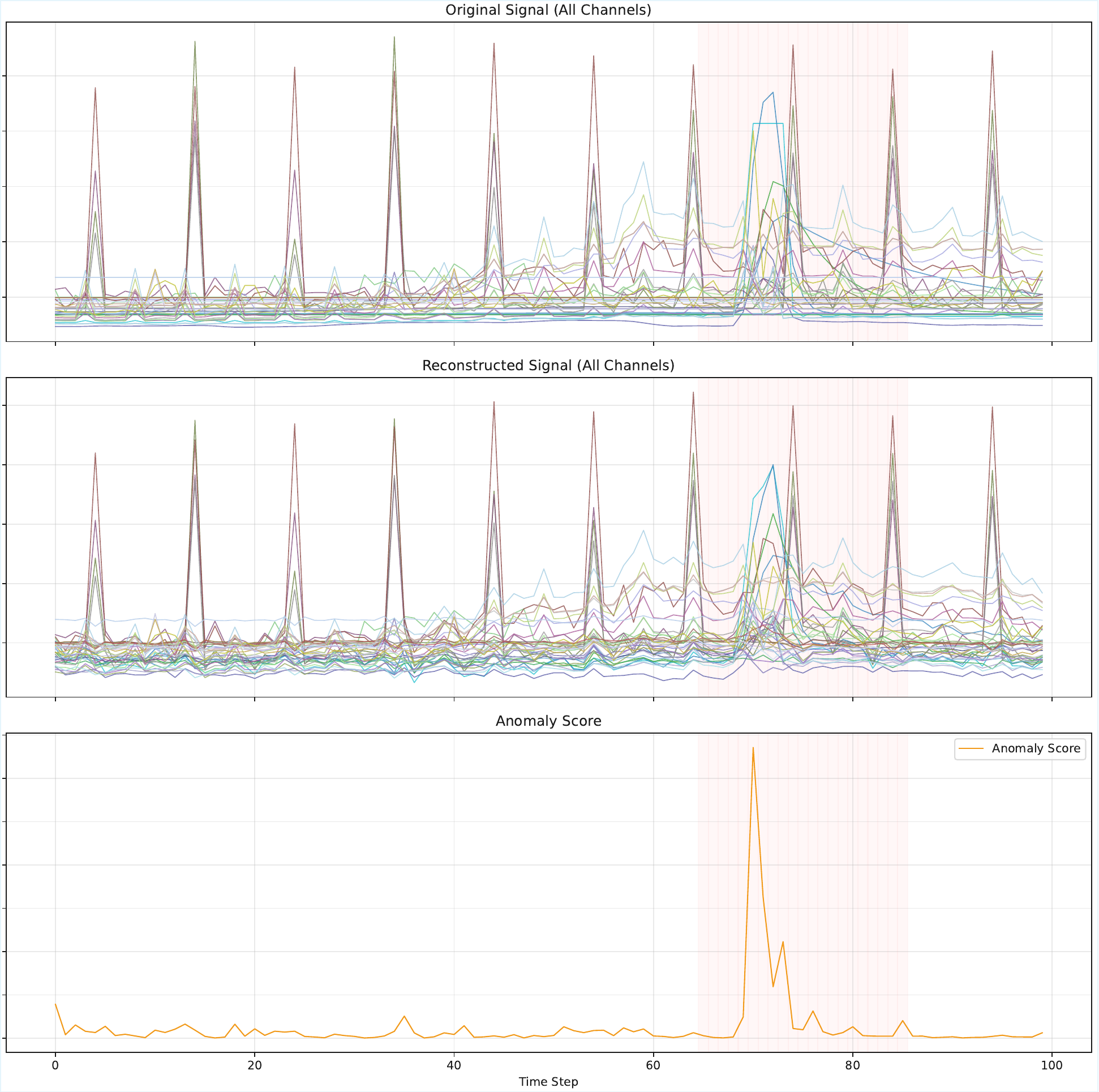}
        &
        \includegraphics[width=0.22\textwidth]{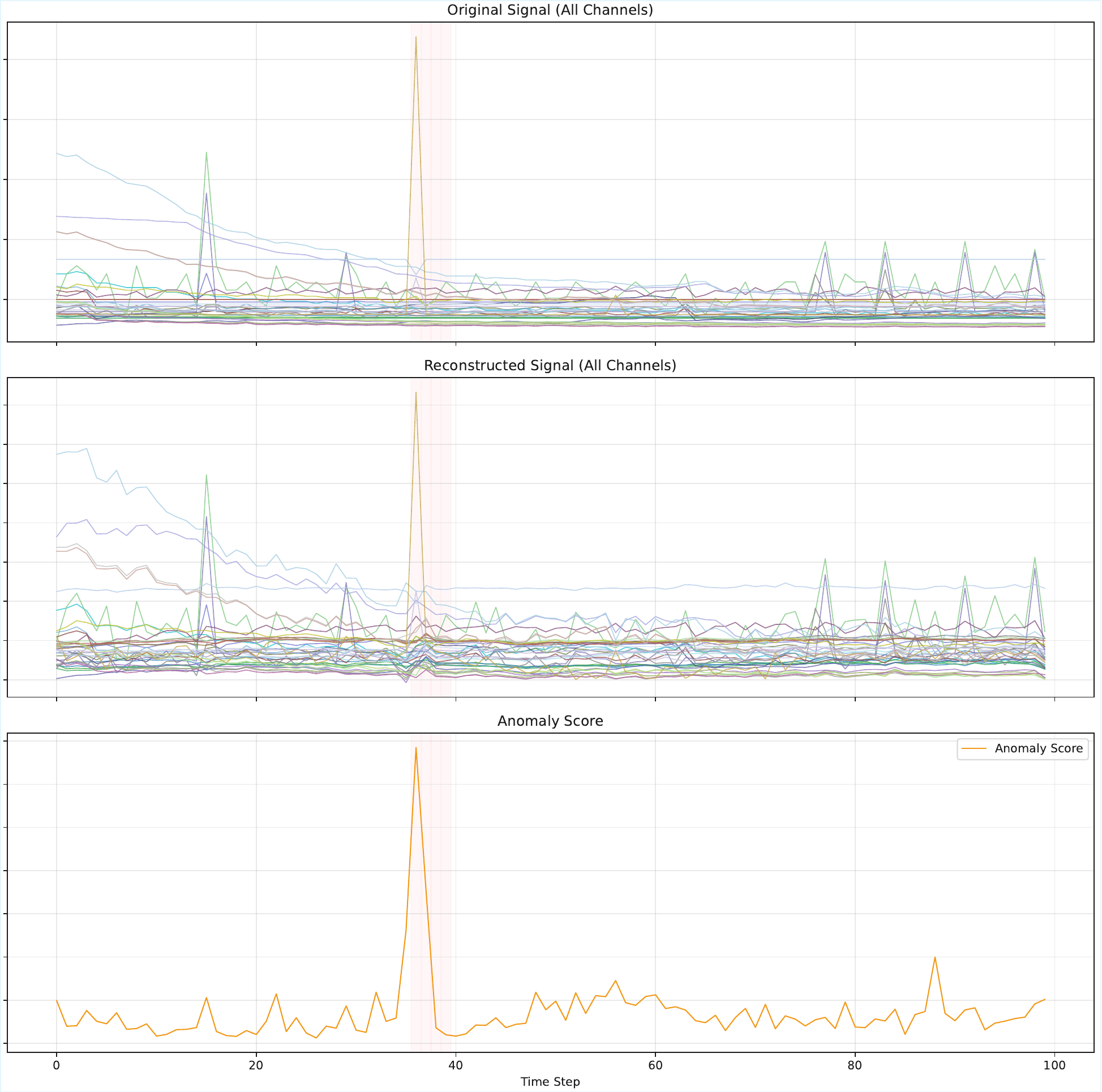}
        &
        \includegraphics[width=0.22\textwidth]{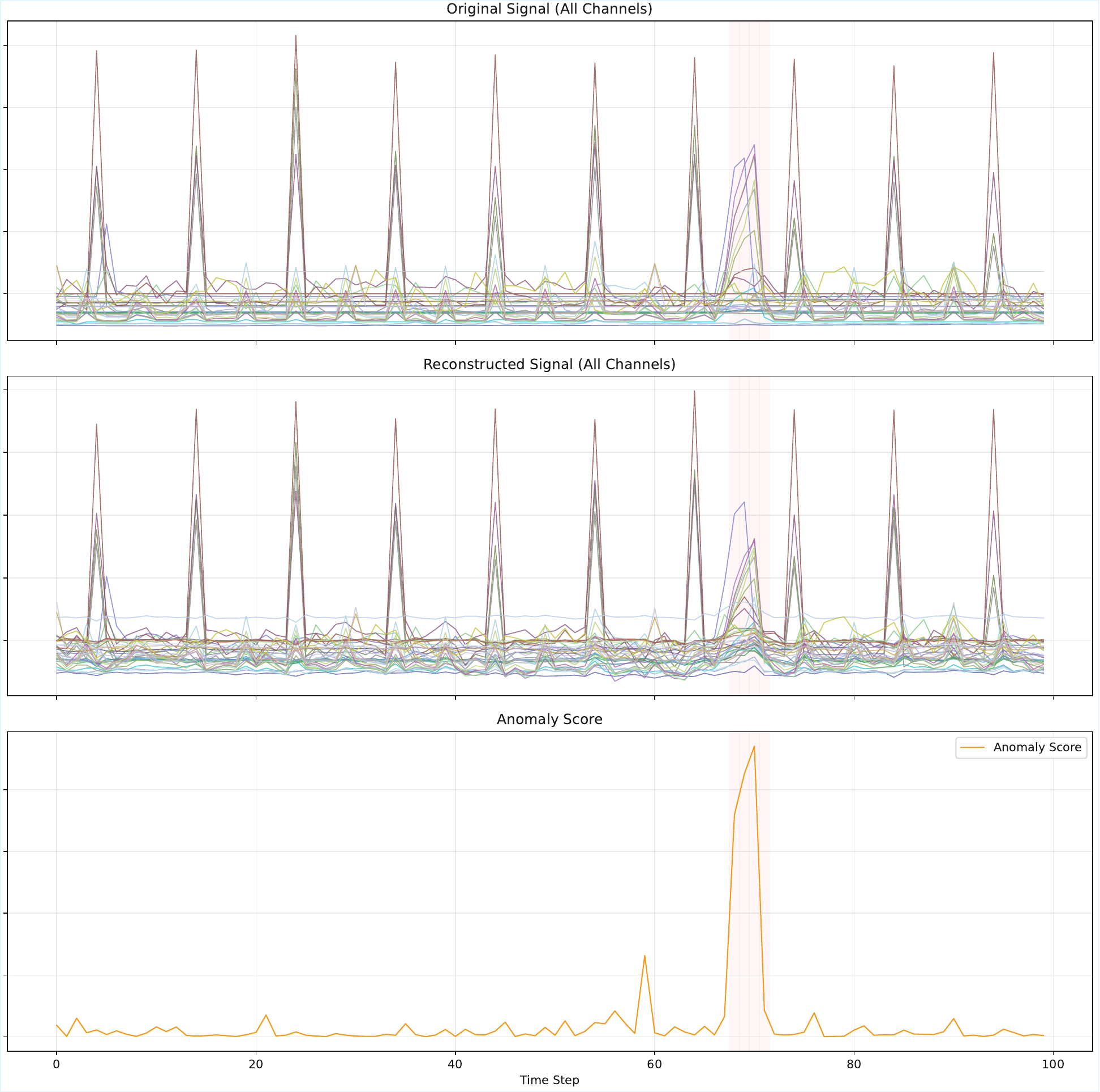}
        &
        \includegraphics[width=0.22\textwidth]{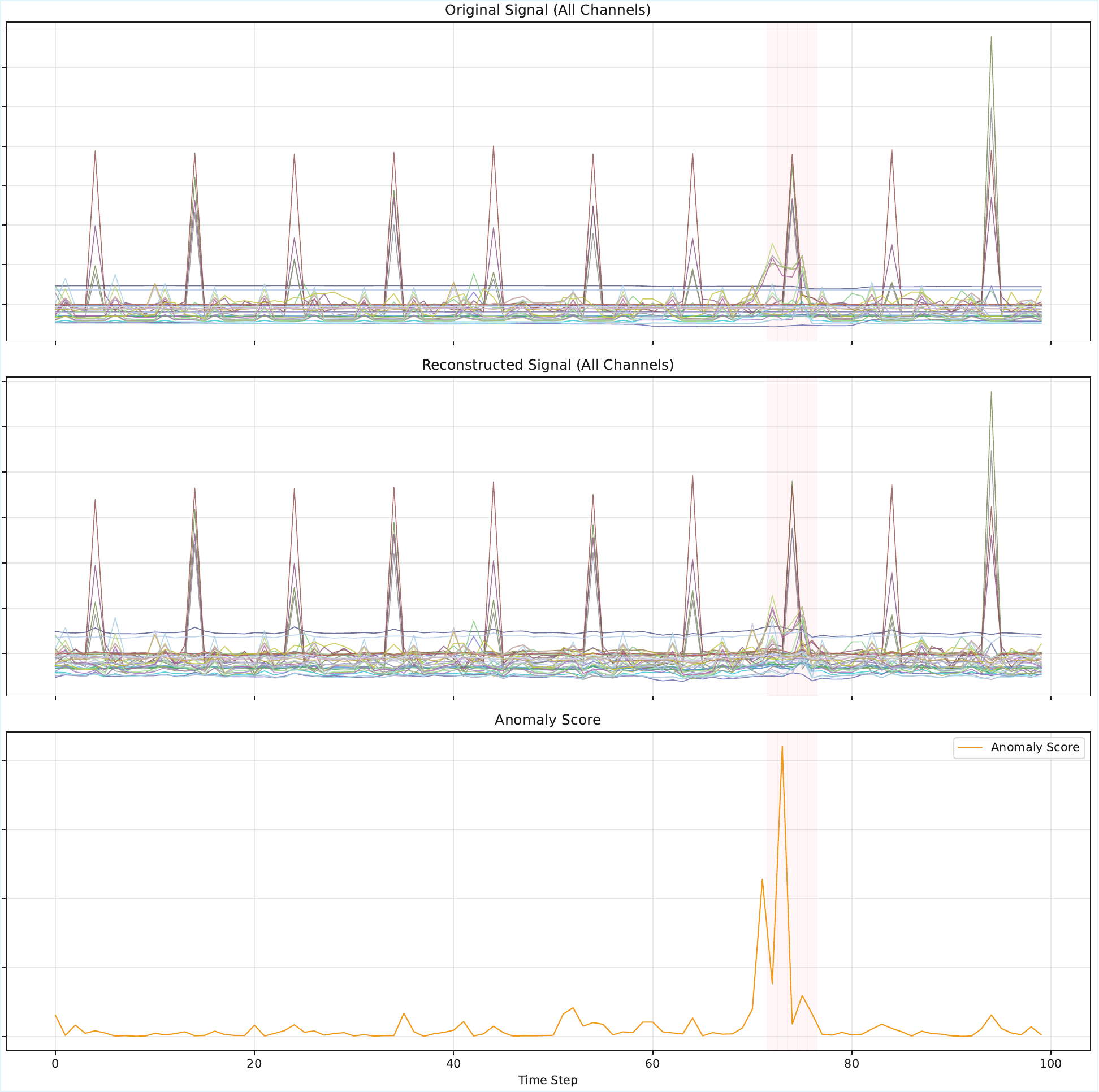}
        \\[6pt]

        % 第 2 行
        \includegraphics[width=0.22\textwidth]{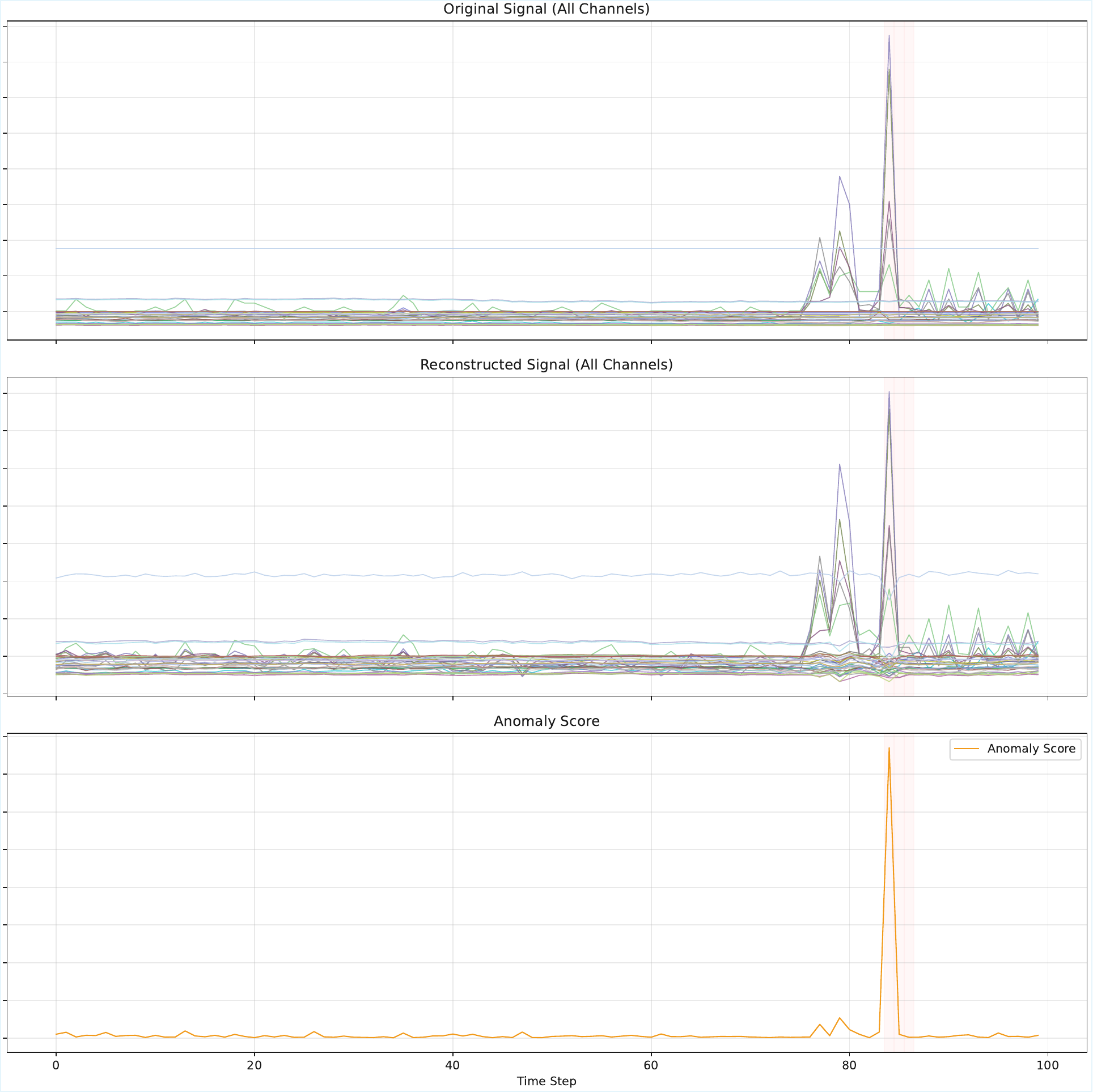}
        &
        \includegraphics[width=0.22\textwidth]{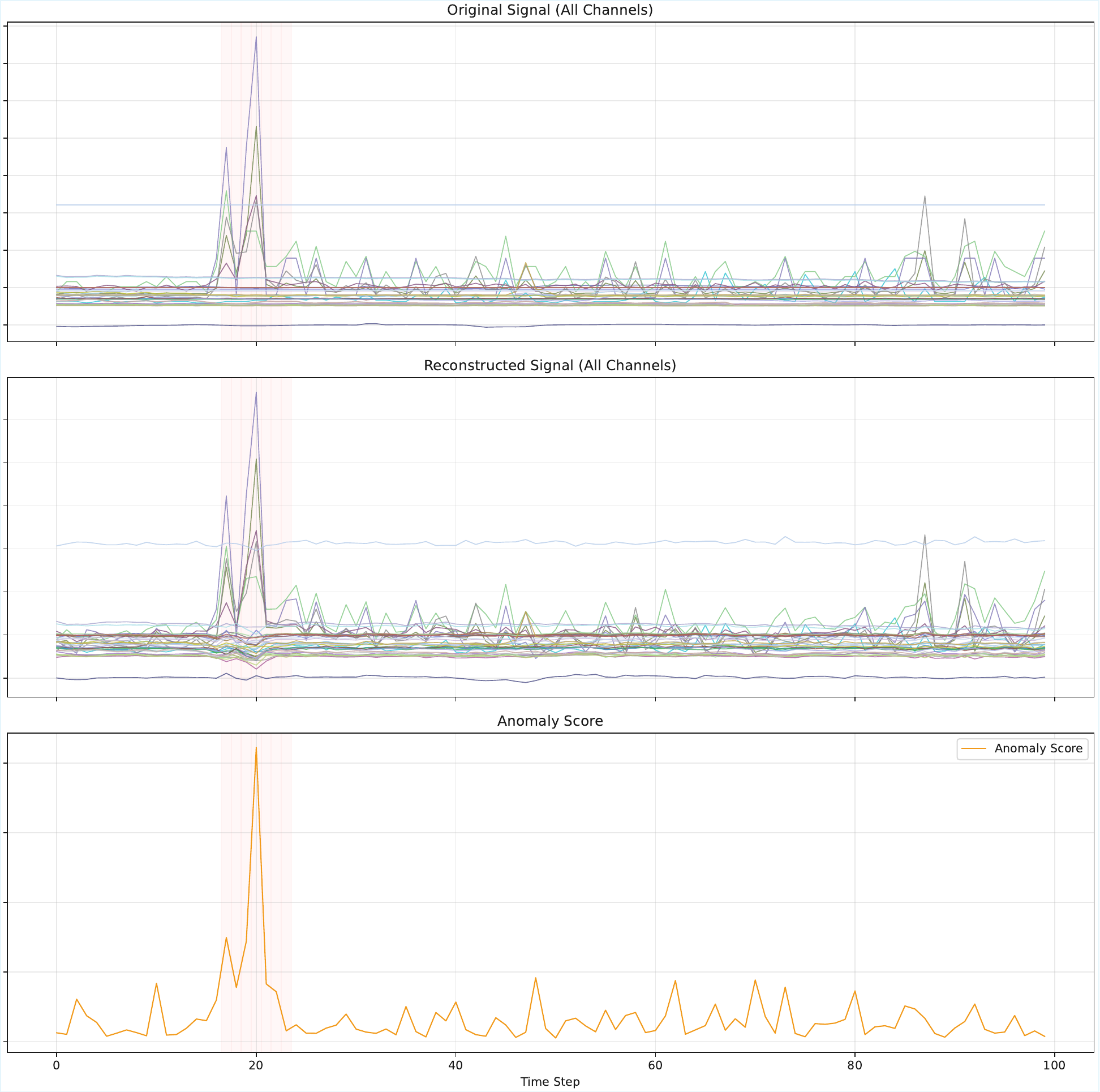}
        &
        \includegraphics[width=0.22\textwidth]{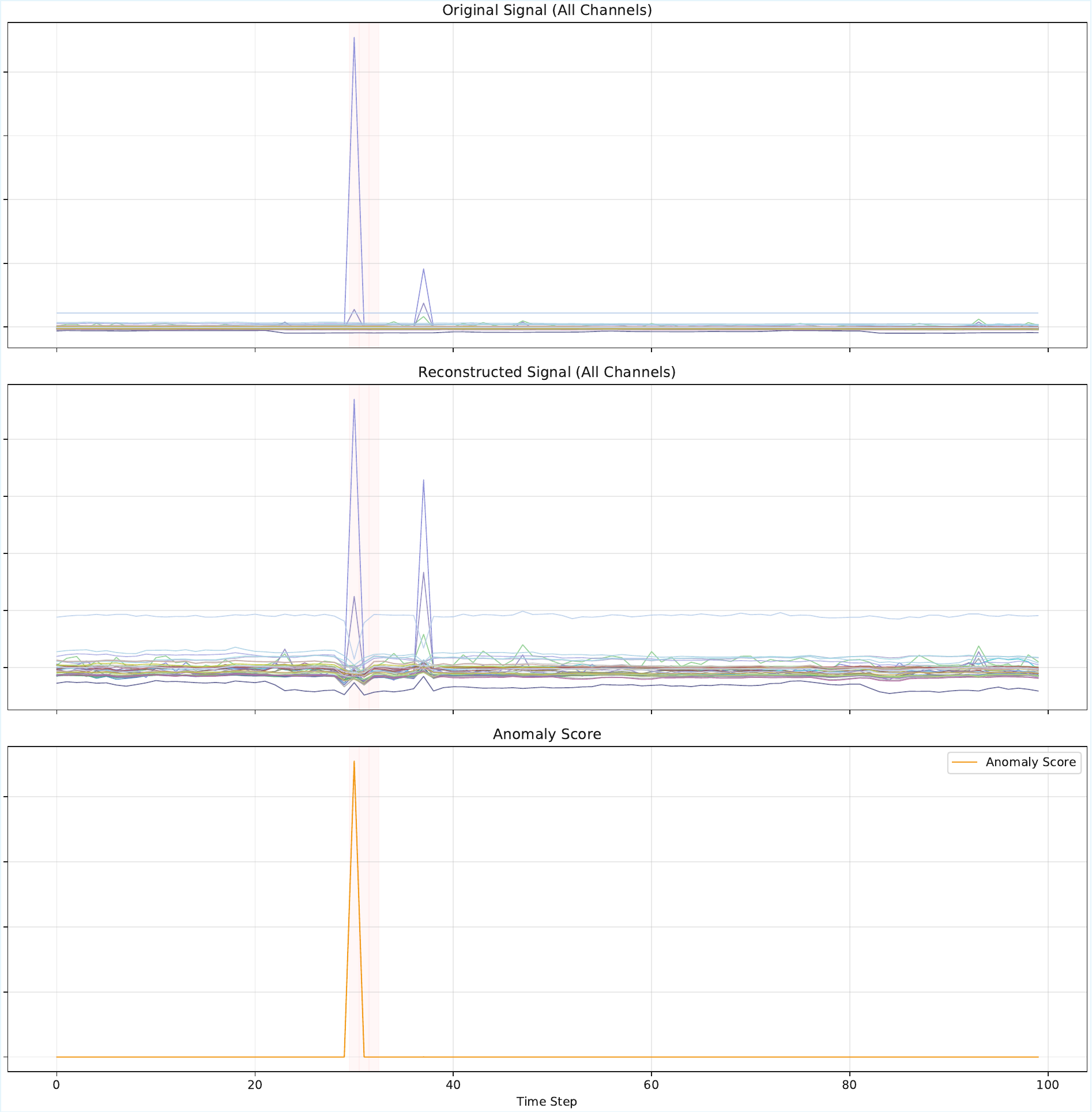}
        &
        \includegraphics[width=0.22\textwidth]{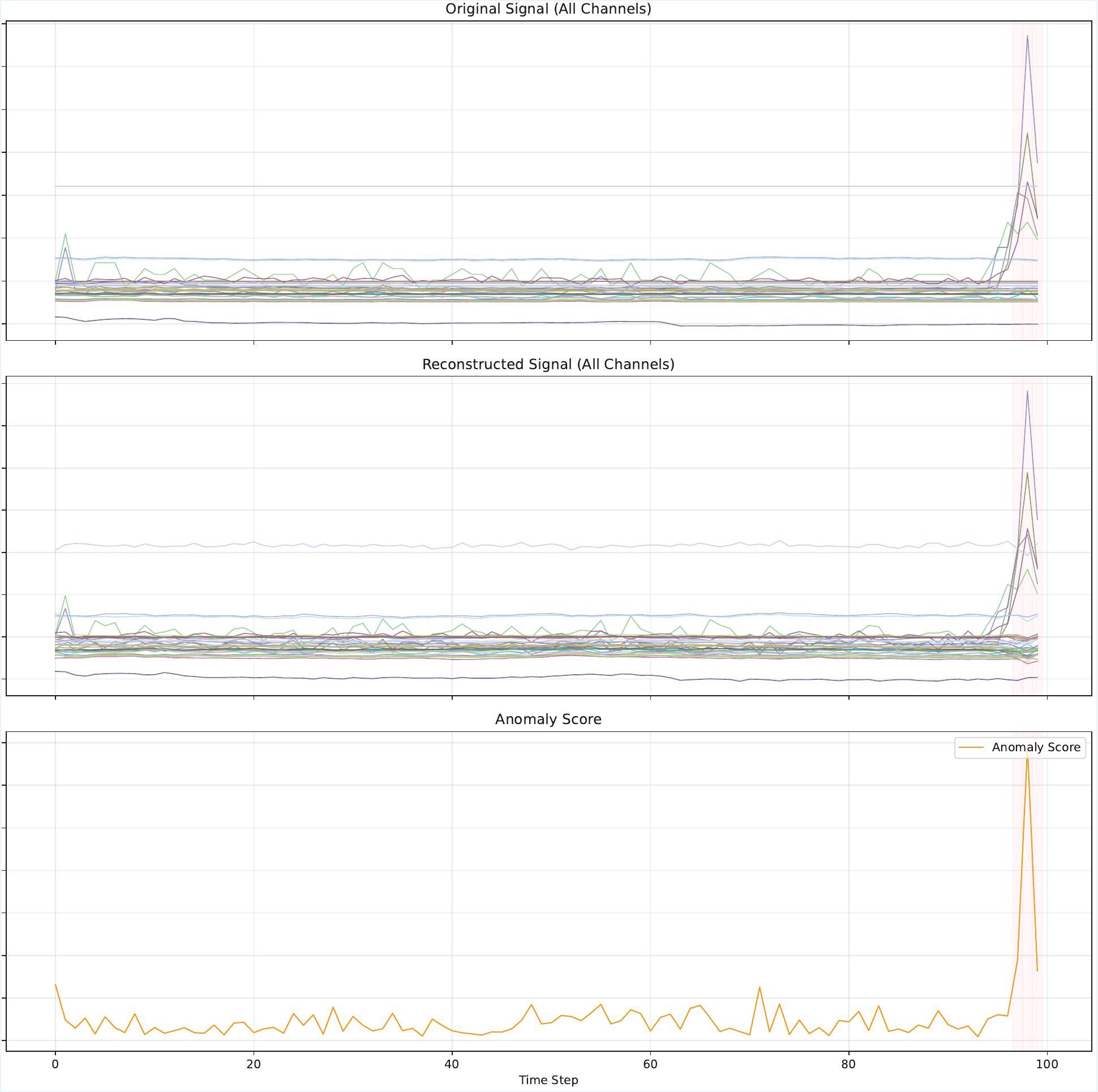}
        \\[6pt]

        % 第 3 行
        \includegraphics[width=0.22\textwidth]{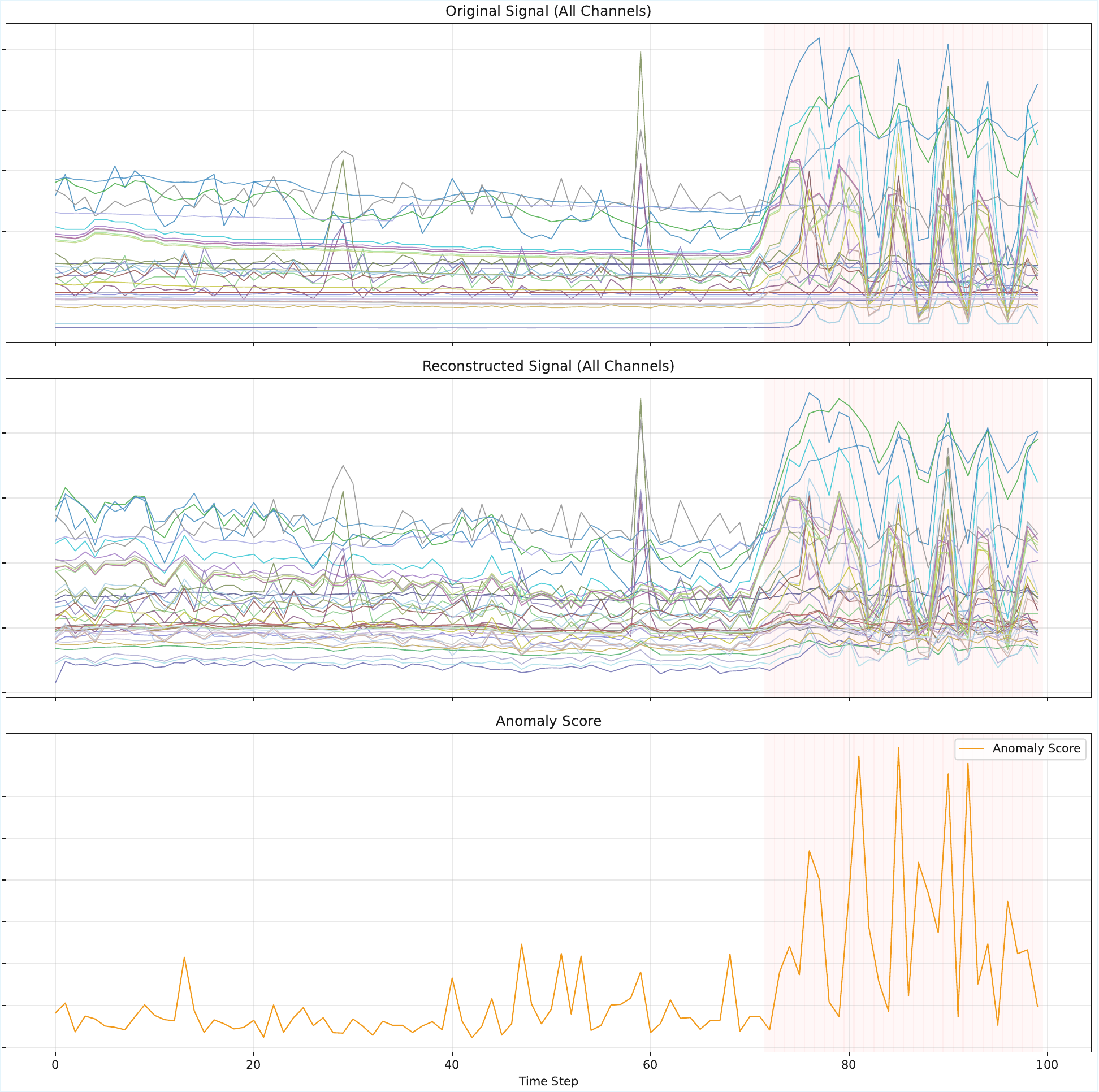}
        &
        \includegraphics[width=0.22\textwidth]{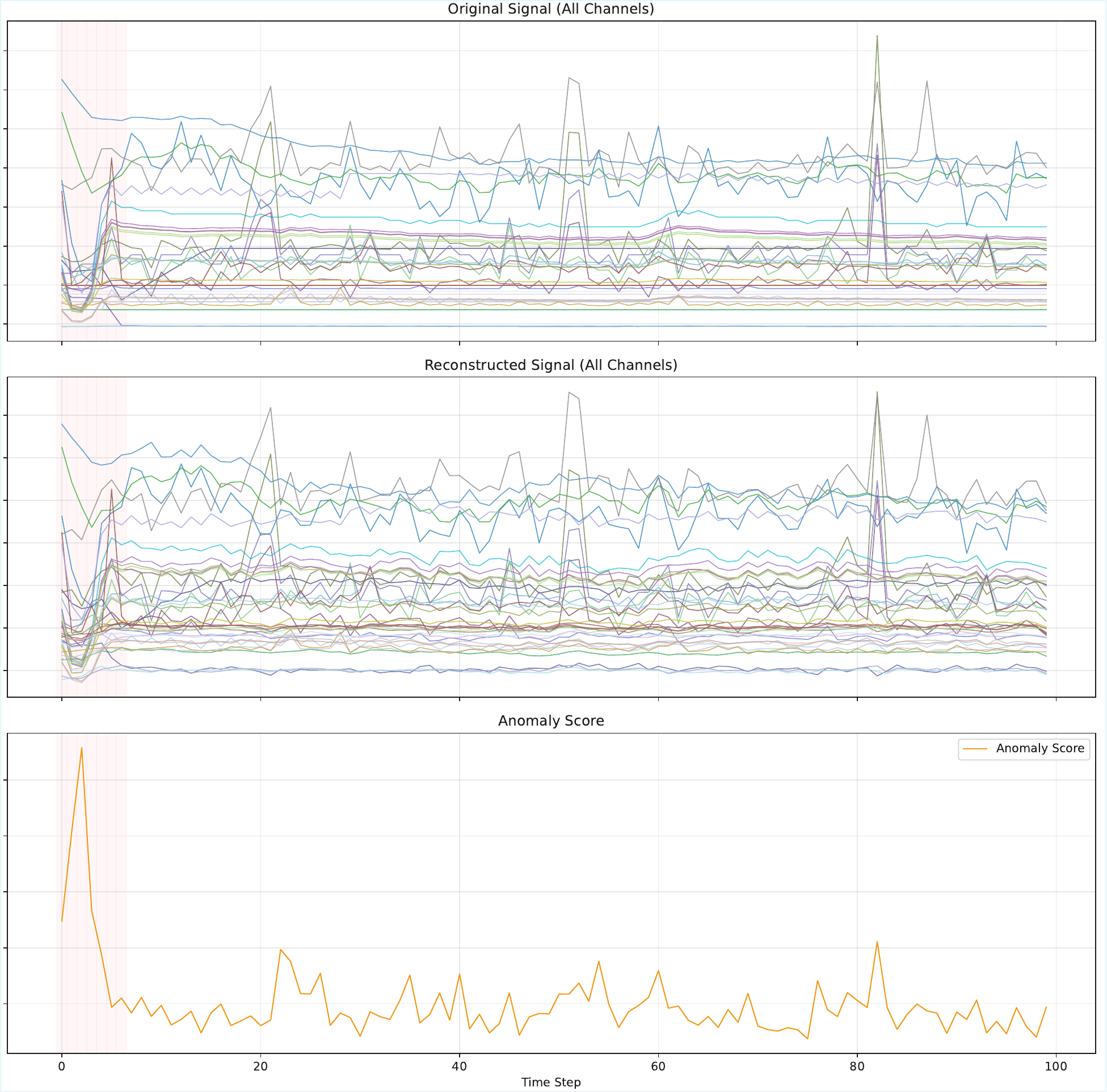}
        &
        \includegraphics[width=0.22\textwidth]{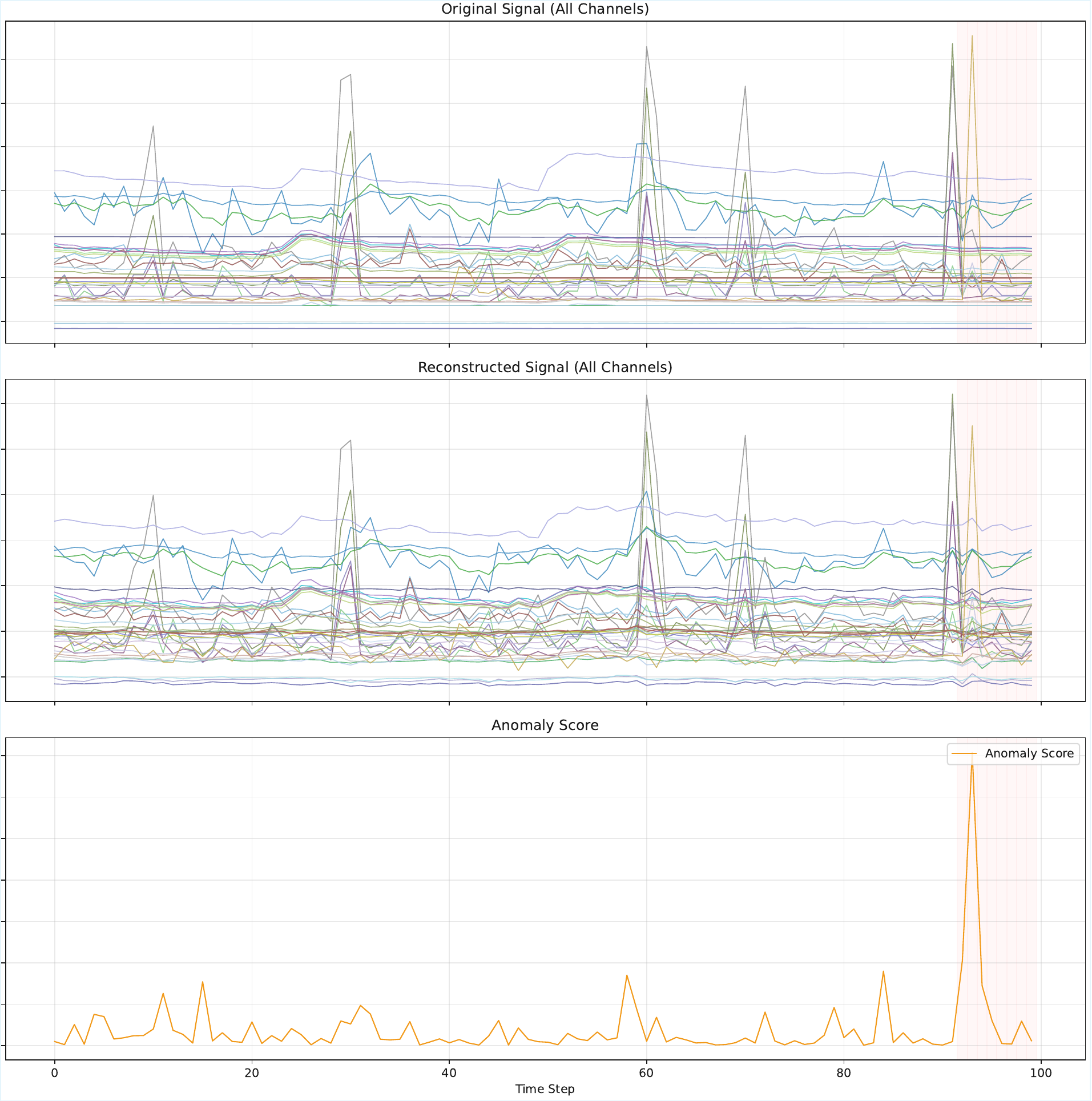}
        &
        \includegraphics[width=0.22\textwidth]{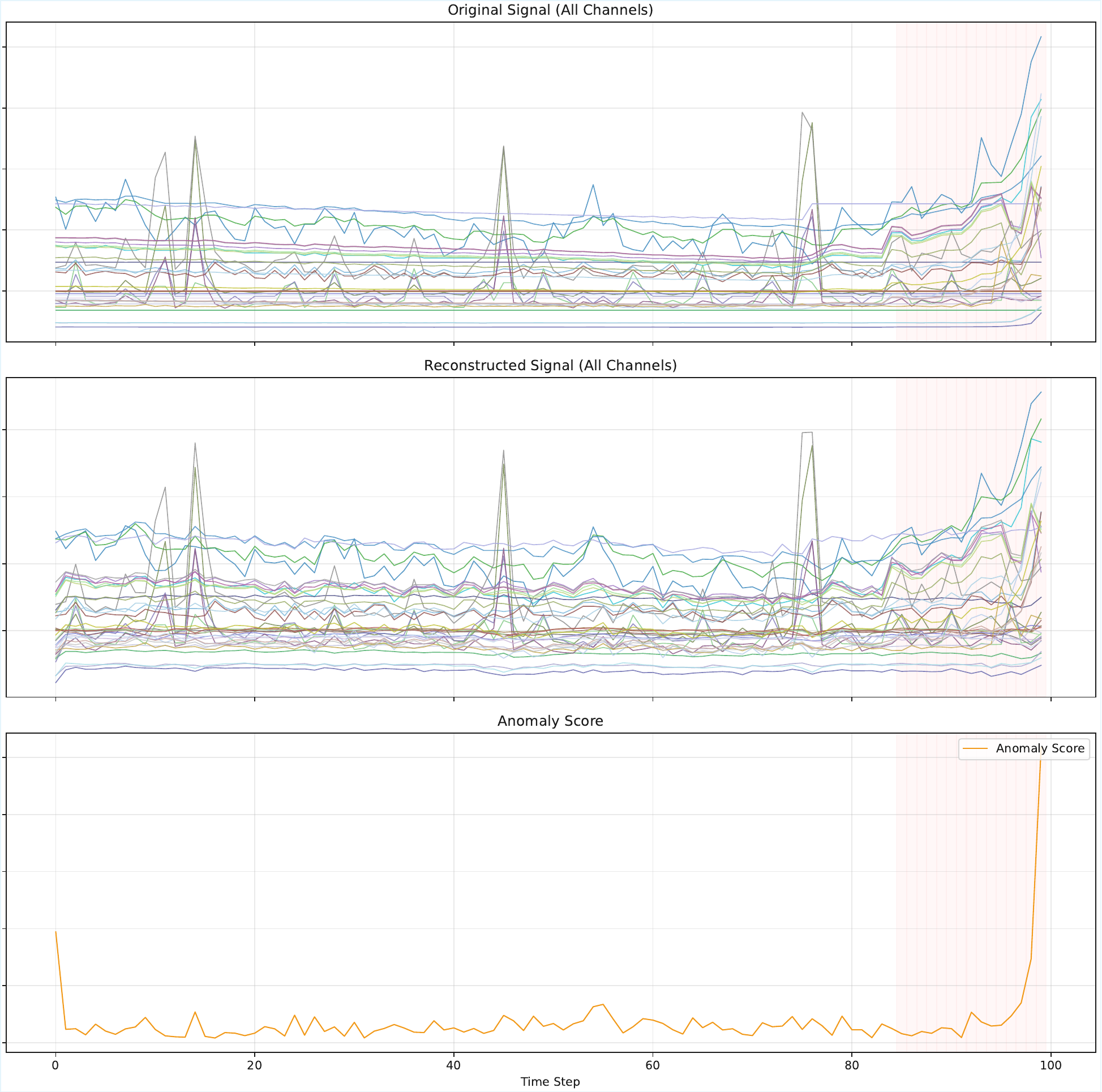}
    \end{tabular}
    \vspace{0pt}
    \caption{Visualization of more anomaly detection results. From top to bottom in each subplot: time series signal, reconstruction signal, anomaly score.}
    \label{fig:vis_all}
\end{figure*}

\begin{table}[htbp]
  \centering
  \caption{A comprehensive evaluation of DBR-AF versus CATCH using additional metrics across five benchmark datasets.}
  \label{tab:add_metric}
    \resizebox{1\linewidth}{!}{
  \begin{tabular}{cccccccccc}
    \toprule
    Dataset & Method & Acc & F1 & Aff-P & Aff-R & R-AR & R-AP & V-ROC & V-PR \\
    \midrule
    \multirow{2}{*}{SMD}       & DBR-AF       & \textbf{99.53} & \textbf{94.80} & 69.12 & 92.99 & \textbf{87.33} & \textbf{28.51} & \textbf{86.41} & \textbf{27.54} \\
                            
                              & CATCH      & 95.01 & 62.08 & \textbf{76.14} & \textbf{93.07} & 84.04 & 24.08 & 83.34 & 23.30 \\
    \midrule
    \multirow{2}{*}{MSL}       & DBR-AF       & \textbf{99.00} & \textbf{95.59} & 57.01 & \textbf{96.68} & \textbf{70.23} & 19.92 & \textbf{70.53} & 19.72 \\
                            
                              & CATCH      & 94.48 & 78.24 & \textbf{57.38} & 96.02 & 68.56 & \textbf{20.90} & 67.81 & \textbf{20.56} \\
    \midrule
    \multirow{2}{*}{SMAP}      & DBR-AF       & \textbf{99.30} & \textbf{96.31} & \textbf{53.11} & \textbf{97.94} & \textbf{64.12} & \textbf{18.04} & \textbf{63.95} & \textbf{18.04} \\
                            
                              & CATCH      & 92.93 & 67.31 & 45.42 & 70.02 & 52.55 & 12.34 & 52.50 & 12.33 \\
    \midrule
    \multirow{2}{*}{SWaT}      & DBR-AF       & \textbf{99.37} & \textbf{97.66} & \textbf{61.03} & \textbf{98.44} &\textbf{81.73}  & \textbf{62.67} & \textbf{81.64} &\textbf{62.61} \\
                            
                              & CATCH      & 96.89 & 87.58 & 57.15 & 84.84 & 48.12 & 27.33 & 48.23 & 27.55 \\
        \midrule
    \multirow{2}{*}{PSM}       & DBR-AF       & \textbf{99.41} & \textbf{98.88} & 62.90 & 86.37 &\textbf{81.70}  &\textbf{63.06}  &\textbf{80.91}  &\textbf{62.70}  \\
                            
                              & CATCH      &98.57 &97.45 &\textbf{72.96} &\textbf{93.68} &69.15 &50.32 &68.23 &49.29 \\
    \midrule
    \multirow{2}{*}{Average}   & DBR-AF       & \textbf{99.32} & \textbf{96.65} & 60.63 & \textbf{94.48} &\textbf{77.02}  &\textbf{38.44}  &\textbf{76.69}  &\textbf{38.12}  \\
                            
                              & CATCH      &95.56 &78.53 &\textbf{61.81} &87.53 &64.48 &26.99 &64.02 &26.61 \\
    \bottomrule
  \end{tabular}
    }
\end{table}

\subsection{Multiple Metric Comparison}
In addition to the five evaluation metrics adopted in the main text, we also introduce a more comprehensive metric suite (as presented in \cref{tab:add_metric}) to conduct a rigorous assessment of the performance of our proposed DBR-AF while performing a comparative analysis against CATCH, a state-of-the-art approach in the field. Specifically, Acc stands for Accuracy; Affiliation Precision (Aff-P) and Affiliation Recall (Aff-R) quantify the alignment accuracy by calculating the spatial discrepancies between the predicted events and the ground-truth annotations~\cite{huet2022local}. The range-based evaluation metrics consist of Range-AUC-ROC (R-AR) and Range-AUC-PR (R-AP), both of which are derived via label transformation techniques~\cite{paparrizos2022tsb}. Additionally, Volume Under the Surface (VUS) metrics have attracted considerable attention from researchers in recent years. The specific metrics involved in this paper include VUS-ROC (V-ROC) and VUS-PR (V-PR)~\cite{paparrizos2022tsb}. The results of the comparative analysis demonstrate that our DBR-AF maintains strong competitiveness across multiple evaluation metrics, fully verifying its outstanding robustness and comprehensive performance.

\subsection{Visualization Results}
As shown in ~\cref{fig:vis_all}, we present some visualizations of anomaly detection results, including reconstruction outputs and anomaly scores.

\section{Conclusion}
Our DBR-AF addresses two critical challenges in MTSAD: reconstruction errors from spurious cross-variable correlations and anomaly score distortion due to large reconstruction errors. Extensive experiments on multiple benchmark datasets show our method achieves highly competitive performance on both threshold-dependent and threshold-independent anomaly detection metrics. While effective, our approach lacks sufficient exploration in anomaly detection interpretability. For future work, we can constrain residual transformation directions via customized flow matching to further improve model performance.
 % argument is your BibTeX string definitions and bibliography database(s)
%\bibliography{IEEEabrv,../bib/paper}
%
% \section{Simple References}
% You can manually copy in the resultant .bbl file and set second argument of $\backslash${\tt{begin}} to the number of references
%  (used to reserve space for the reference number labels box).

% \begin{thebibliography}{1}
% \bibliographystyle{IEEEtran}

% \end{thebibliography}
\bibliography{ref}
\bibliographystyle{IEEEtran}

\end{document}